\documentclass[letterpaper]{article} % DO NOT CHANGE THIS
\usepackage{aaai24}  % DO NOT CHANGE THIS
\usepackage{times}  % DO NOT CHANGE THIS
\usepackage{helvet}  % DO NOT CHANGE THIS
\usepackage{courier}  % DO NOT CHANGE THIS
\usepackage[hyphens]{url}  % DO NOT CHANGE THIS
\usepackage{graphicx} % DO NOT CHANGE THIS
\usepackage{natbib}  % DO NOT CHANGE THIS AND DO NOT ADD ANY OPTIONS TO IT
\usepackage{caption} % DO NOT CHANGE THIS AND DO NOT ADD ANY OPTIONS TO IT
\usepackage[table, dvipsnames]{xcolor}

\usepackage{algorithm}
\usepackage{algorithmic}

\usepackage{newfloat}
\usepackage{listings}
\DeclareCaptionStyle{ruled}{labelfont=normalfont,labelsep=colon,strut=off} % DO NOT CHANGE THIS
\floatstyle{ruled}
\newfloat{listing}{tb}{lst}{}
\floatname{listing}{Listing}
\title{Object-Aware Domain Generalization for Object Detection}
\author{
    Wooju Lee\equalcontrib,
    Dasol Hong\equalcontrib, 
    Hyungtae Lim\textsuperscript{\rm $\dag$} and
    Hyun Myung\thanks{Corresponding authors: Dr. Hyungtae Lim and Prof. Hyun Myung}
}
\affiliations{
    Urban Robotics Lab, School of Electrical Engineering, \\Korea Advanced Institute of Science and Technology, Republic of Korea\\
     {\tt\small\{dnwn24, ds.hong, shapelim, hmyung\}@kaist.ac.kr}
}

\usepackage{verbatim}
\usepackage{amssymb}
\usepackage{caption}
\usepackage{subcaption} % added by dshong
\usepackage[labelformat=simple]{subcaption}

\usepackage{bm}
\usepackage[export]{adjustbox}

\usepackage[utf8]{inputenc}
\usepackage{dcolumn} % dcolumn to line up decimals
\usepackage{multirow}
\usepackage{adjustbox}
\usepackage{booktabs,caption}
\usepackage[flushleft]{threeparttable} 
\newcommand\mc[1]{\multicolumn{1}{c}{#1}} % handy shortcut macro
\usepackage{pifont}
\newcommand{\xmark}{\ding{55}}%

\def\figref#1{Figure~\ref{#1}}
\def\tabref#1{Table~\ref{#1}}
\def\eqref#1{(\ref{#1})}

\usepackage{pgf}
\usepackage{layouts}
\usepackage{import}
\usepackage{pdfpages}
\usepackage{svg}

\usepackage{makecell}
\usepackage{amsmath}
\usepackage{mathtools}

\begin{document}

\maketitle

\begin{abstract}
Single-domain generalization (S-DG) aims to generalize a model to unseen environments with a single-source domain. However, most S-DG approaches have been conducted in the field of classification. When these approaches are applied to object detection, the semantic features of some objects can be damaged, which can lead to imprecise object localization and misclassification.
To address these problems, we propose an object-aware domain generalization (OA-DG) method for single-domain generalization in object detection. Our method consists of data augmentation and training strategy, which are called \textit{OA-Mix} and \textit{OA-Loss}, respectively. OA-Mix generates multi-domain data with multi-level transformation and object-aware mixing strategy. OA-Loss enables models to learn domain-invariant representations for objects and backgrounds from the original and OA-Mixed images. Our proposed method outperforms state-of-the-art works on standard benchmarks. Our code is available at https://github.com/WoojuLee24/OA-DG.
%Our code will be made available on publication.
\end{abstract}

\section{Introduction}
Modern deep neural networks (DNNs) have achieved human-level performances in various applications such as image classification and object detection~\cite{he2016deep, dosovitskiy2021an, carion2020end, ren2015faster}. However, DNNs are vulnerable to various types of domain shifts, which have not been seen in the source domain~\cite{hendrycks2018benchmarking, lee2022adversarial, michaelis2019benchmarking, wu2022single}. Even a small change in the domain can have disastrous results in real-world scenarios such as autonomous driving~\cite{michaelis2019benchmarking, wu2022single}. Thus, DNNs should be robust against the domain shifts to be applied in real-world applications.

Domain generalization (DG) aims to generalize a model to unseen target domains by using only the source domain~\cite{hendrycksaugmix, kim2021selfreg, yao2022pcl, zhou2022domain}. However, most DG methods rely on multiple source domains and domain annotations, which are generally unavailable~\cite{kim2021selfreg, lin2021domain, yao2022pcl}. Single-domain generalization (S-DG) achieves DG without any additional knowledge about multiple domains~\cite{wan2022meta, wang2021learning}.

\begin{figure}[!t]
     \begin{center}
         \begin{subfigure}{0.46\textwidth}
             \centering
             \includegraphics[width=\textwidth]{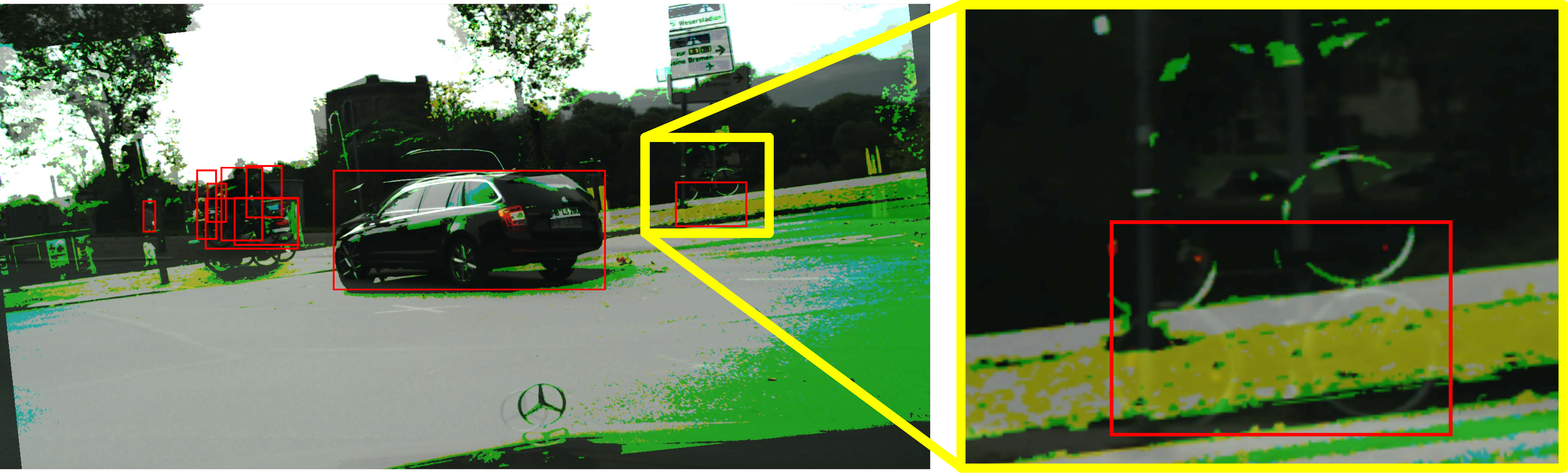}
             \caption{AugMix~\shortcite{hendrycksaugmix}}
             \label{fig:problem__data_augmentation}
         \end{subfigure}
         \begin{subfigure}{0.46\textwidth}
             \centering
             \includegraphics[width=\textwidth]{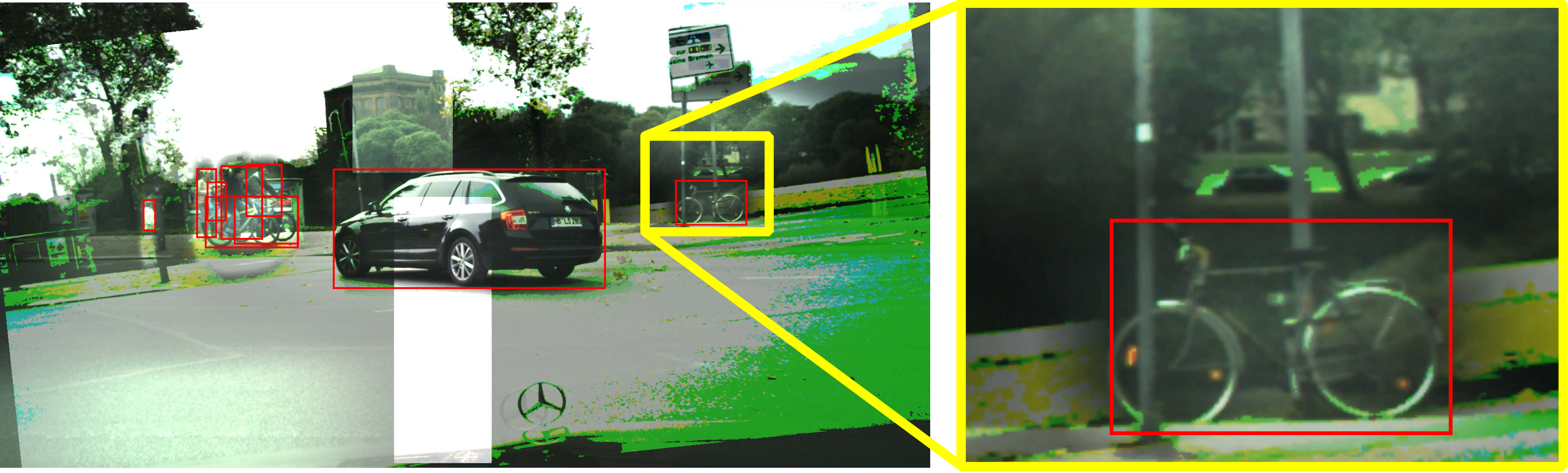}
             \caption{OA-Mix}
             \label{fig:solution__data_augmentation}
         \end{subfigure}
         \begin{subfigure}{0.153\textwidth}
             \centering
             \includegraphics[width=\textwidth]{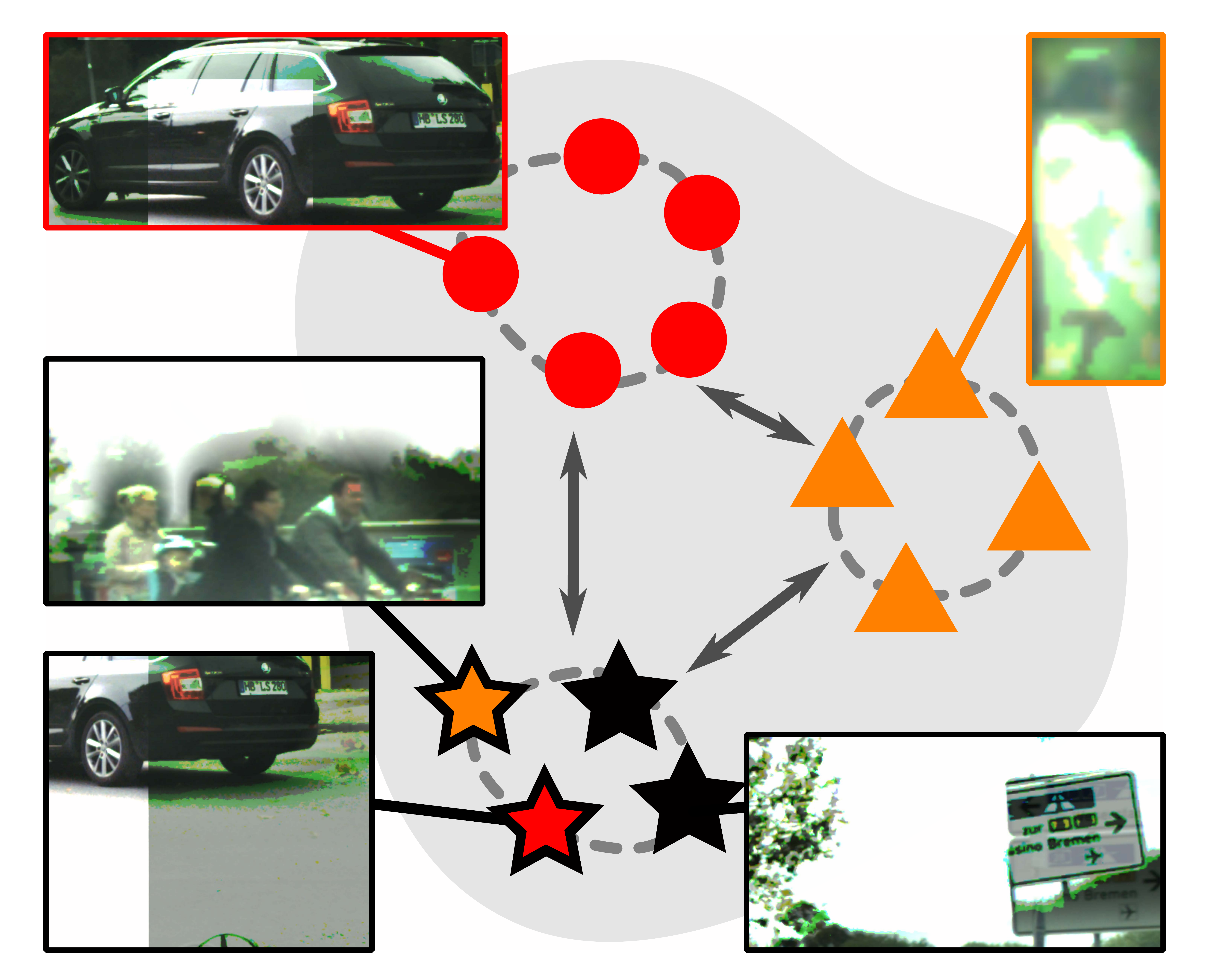}
             \caption{SupCon~\shortcite{khosla2020supervised}}
             \label{fig:comparision_contrastive__supcon}
         \end{subfigure}
         \begin{subfigure}{0.153\textwidth}
             \centering
             \includegraphics[width=\textwidth]{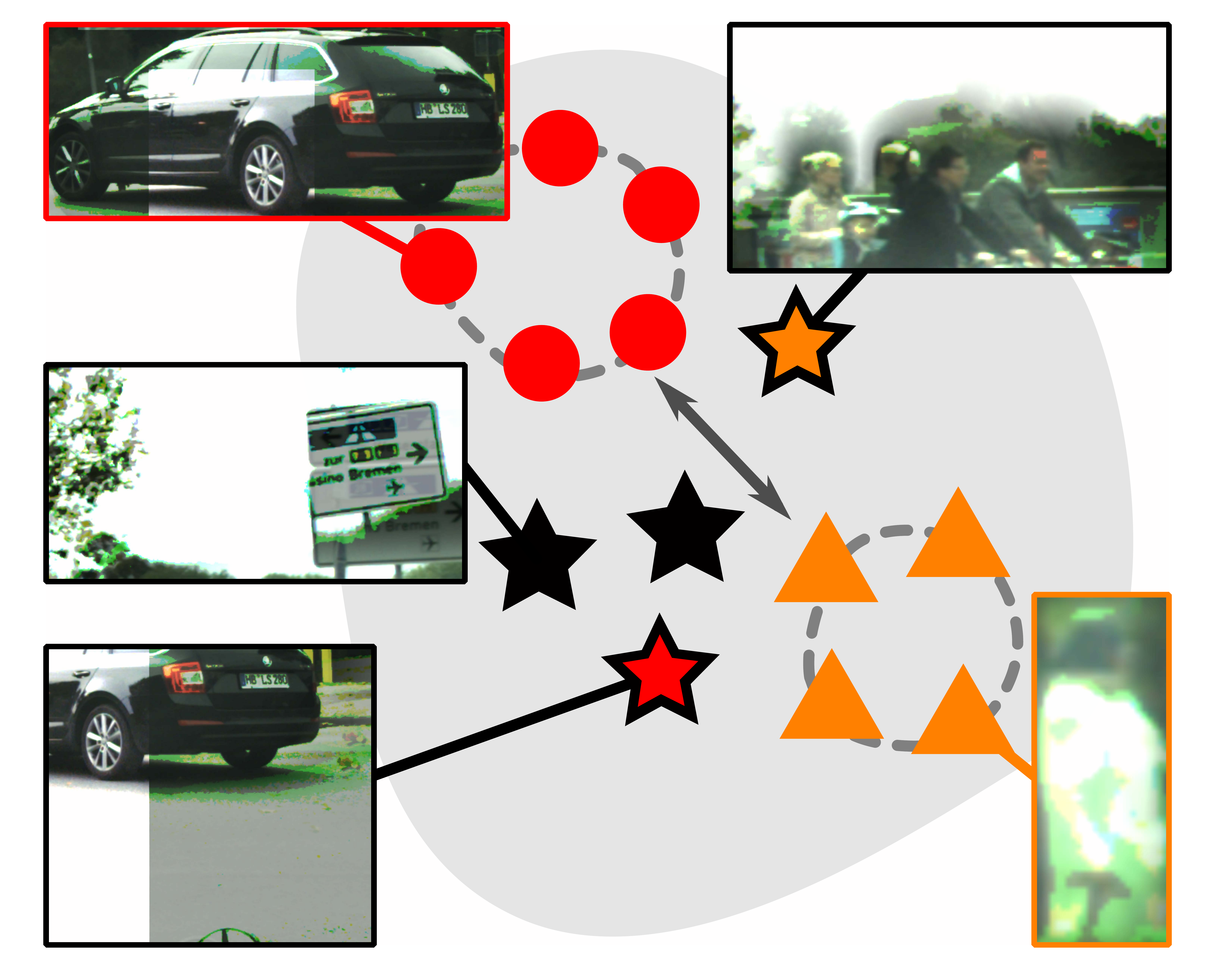}
             \caption{FSCE~\shortcite{sun2021fsce}}
             \label{fig:comparision_contrastive__fsce}
         \end{subfigure}
         \begin{subfigure}{0.153\textwidth}
             \centering
             \includegraphics[width=\textwidth]{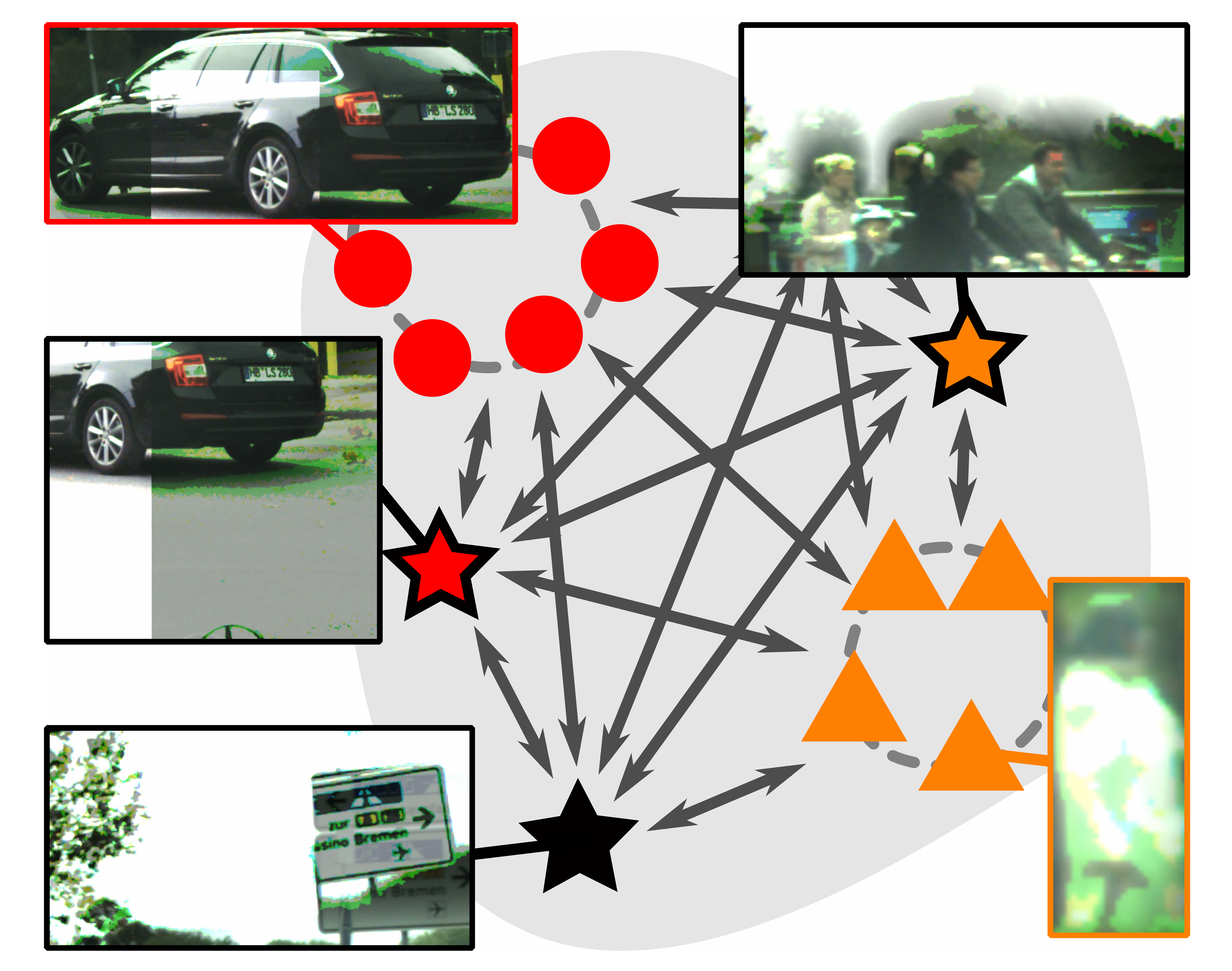}
             \caption{OA-Loss}
             \label{fig:comparision_contrastive__olaloss}
         \end{subfigure}
     \end{center}
     % \caption{
     % (a) Existing data augmentation methods can damage the semantic features of objects. (b) OA-Mix generates multiple domains while preserving semantic features.
     % (c)-(e) The circles, triangles, and stars represent the car, person, and background classes, respectively. Existing methods (c) and (d) ignore the fact that background instances can include parts of different objects. (e) OA-Loss considers semantic relations of background instances from multi-domain. 
     % }
     \caption{
     (a) Existing data augmentation methods can damage the semantic features of objects. (b) OA-Mix generates multiple domains while preserving semantic features.
     (c)-(e) The dotted line, arrow, circle, triangle, and star mean pull, push, car, person, and background, respectively.
     % Existing methods (c) and (d) ignore the fact that background instances can include parts of different objects. 
     (c) The background contains different semantics (red and orange stars), but these are treated identically and pulled. (d) The method ignores semantic relationships between background instances. (e) OA-Loss considers semantic relations of background instances from multi-domain. 
     }
     % (c): The background contains different semantics (red and orange stars), but these are treated identically and pulled. (e): OA-DG distinguishes different semantics of background, pushing them apart
\end{figure}

Data augmentation has been successfully applied for S-DG in image classification~\cite{hendrycksaugmix, modas2022prime}. The methods can be used to generate multi-source domains from a single-source domain. However, object detection addresses multiple objects in an image. When data augmentation methods for S-DG are applied to object detection, object annotations may be damaged.
As shown in~\figref{fig:problem__data_augmentation}, spatial and color transformations can damage the positional or semantic features of objects.
Michaelis~et~al.~\shortcite{michaelis2019benchmarking} avoided this problem with style-transfer augmentations that do not change object locations.
However, this approach does not leverage a rich set of transformations in image classification, thus limiting the domain coverage.
Therefore, data augmentation for single-source domain generalization for object detection (S-DGOD) should include various transformations without damaging object annotation.

Recently, contrastive learning methods have been proposed to reduce the gap between the original and augmented domains~\cite{kim2021selfreg, yao2022pcl}. The methods address inter-class and intra-class relationships from multi-domain. However, they only train the relationships in object classes and do not consider background class for object detection~\cite{sun2021fsce} as shown in~\figref{fig:comparision_contrastive__fsce}.
% The background class is required for object detectors to classify objectness. 
Because the object detector misclassifies foreground as background in OOD, considering the background class is required to classify the presence of objects in OOD.
Therefore, the contrastive learning method should consider both the foreground and background classes for object detectors to classify objectness in out-of-distribution.

In this study, we propose object-aware domain generalization method~(OA-DG) for S-DGOD to overcome the aforementioned limitations. Our method consists of OA-Mix for data augmentation and OA-Loss for reducing the domain gap. OA-Mix consists of multi-level transformations and object-aware mixing.
Multi-level transformations introduce local domain changes within an image and object-aware mixing prevents the transformations from damaging object annotation.
OA-Mix is the first object-aware approach that allows a rich set of image transformations for S-DGOD.

OA-Loss reduces the gap between the original and augmented domains in an object-aware manner. 
For foreground instances, the method trains inter-class and intra-class relations to improve object classification in out-of-distribution. Meanwhile, the background instances that belong to the same class can partially include the foreground objects of different classes.
OA-Loss allows the model to learn the semantic relations among background instances in multi-domain as illustrated in~\figref{fig:comparision_contrastive__olaloss}. To the best of our knowledge, the proposed method first trains the semantic relations for both the foreground and background instances to achieve DG with contrastive methods.

Our proposed method shows the best performance on common corruption~\cite{michaelis2019benchmarking} and various weather benchmarks~\cite{ wu2022single} in an urban scene. The contributions can be summarized as follows.

\begin{itemize}
    \item We propose OA-Mix, a general and effective data augmentation method for S-DGOD. It increases image diversity while preserving important semantic features with multi-level transformations and object-aware mixing.
    \item We propose OA-Loss that reduces the domain gap between the original and augmented images. OA-Loss enables the model to learn semantic relations of foreground and background instances from multi-domain.
    \item Extensive experiments on standard benchmarks show that the proposed method outperforms state-of-the-art methods on unseen target domains.
\end{itemize}

%========================================================================
\section{Related work}
%-------------------------------------------------------------------------
\subsection{Data augmentation for domain generalization}
Recently, many successful data augmentation methods for S-DG have been proposed for image classification. AugMix~\cite{hendrycksaugmix} mixes the results of the augmentation chains to generate more diverse domains. Modas~et~al.~\shortcite{modas2022prime} generated diverse domains using three primitive transformations with smoothness and strength parameters. However, direct application of these methods to object detection can damage the locations or semantic annotations of objects, as shown in~\figref{fig:problem__data_augmentation}.
Geirhos~et~al.~\shortcite{geirhosimagenet} avoided this issue by using style transfer~\cite{gatys2016image} that does not damage object annotations and improved robustness to synthetic corruptions.
However, data augmentation methods for S-DGOD should generate diverse domains with a broad range of transformations rather than a limited set of transformations.
The proposed OA-Mix uses an object-aware approach to create various domains without damaging object annotations.

\begin{figure*}[t]
    \begin{center}
        \includegraphics[width=1.0\linewidth]{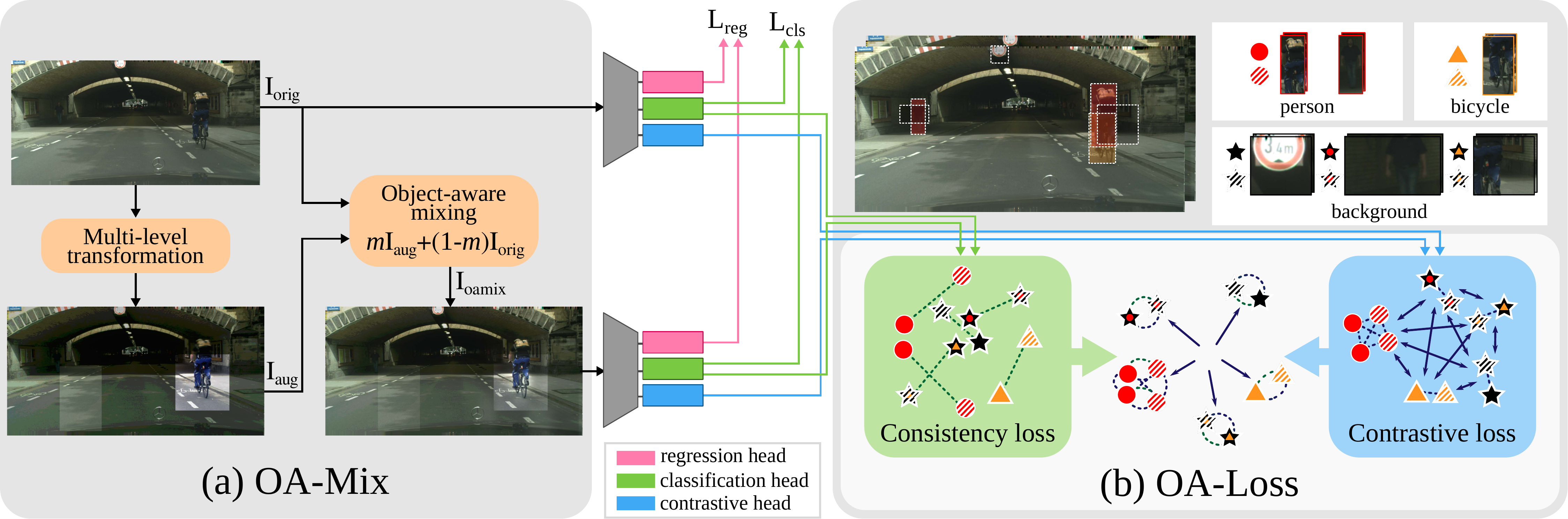}
    \end{center}
    \caption{
            Overview of OA-DG method. 
            (a) OA-Mix transforms an image at multiple levels and mixes the transformed image with the original image in an object-aware manner. The original and augmented images are fed into the object detectors that share weights.
            (b) OA-Loss trains the object detectors to learn the domain-invariant representations from the original and augmented images. Circles, triangles, and stars represent car, person, and background classes, respectively. An augmented instance has white slanted lines. Another shape within a star indicates the partially included object in the background. The consistency loss aligns the original and augmented instances (dotted line). The contrastive loss aligns the positive pairs (dotted line) and repulses the negative pairs (arrow).
        }
        \label{fig:the overview of OA}
\end{figure*}

\subsection{Contrastive learning for domain generalization}
The contrastive learning methods for domain generalization build domain-invariant representations from multi-domain with sample-to-sample pairs~\cite{yao2022pcl, kim2021selfreg, li2021progressive}. The methods pull positive sample pairs of multi-domain and push negative samples away. PCL~\cite{yao2022pcl} arranges representative features and sample-to-sample pairs from multi-domain. SelfReg~\cite{kim2021selfreg} aligns positive sample pairs and regularizes features with a self-supervised contrastive manner. PDEN~\cite{li2021progressive} progressively generates diverse domains and arranges features with a contrastive loss. However, these contrastive learning methods address semantic relations of sample-to-sample pairs only in object~classes, not the background class.

%-------------------------------------------------------------------------

\subsection{Single-domain generalization for object detection}
Recently, several S-DGOD methods have been proposed. CycConf~\cite{wang2021robust} enforces the object detector to learn invariant representations across the same instances under various conditions. However, CycConf requires annotated video datasets, which are not generally given.
CDSD~\cite{wu2022single} propagates domain-invariant features to the detector with self-knowledge distillation. 
However, the model does not diversify the single-domain with data augmentation, which potentially can be improved. Vidit~et~al. ~\shortcite{vidit2023clip} utilized a pre-trained vision-language model to generalize a detector, but textual hints about the target domains should be given. Our proposed method does not require prior knowledge about target domains and achieves S-DGOD in an object-aware approach.

%========================================================================
\section{Method}
We propose OA-DG method for S-DGOD. Our approach consists of OA-Mix, which augments images, and OA-Loss, which achieves DG based on these augmented images. OA-DG can be applied to both one-stage and two-stage object detectors. The overview of OA-DG is illustrated in~\figref{fig:the overview of OA}.

%------------------------------------------------------------------------
\subsection{OA-Mix}

\begin{figure}[t]
    \begin{center}
    \begin{subfigure}{0.95\columnwidth}
             \includegraphics[width=\columnwidth]{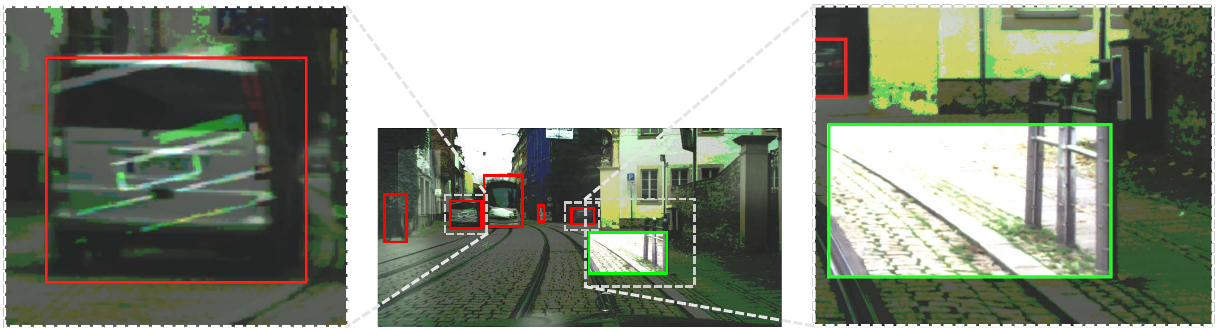} % oamix_multilevel_v3
            \caption{} 
            \label{fig:problem__data_augmentation a}
    \vspace{0.3\baselineskip}
    \end{subfigure}
    \begin{subfigure}{0.95\columnwidth}
        \includegraphics[width=\columnwidth]{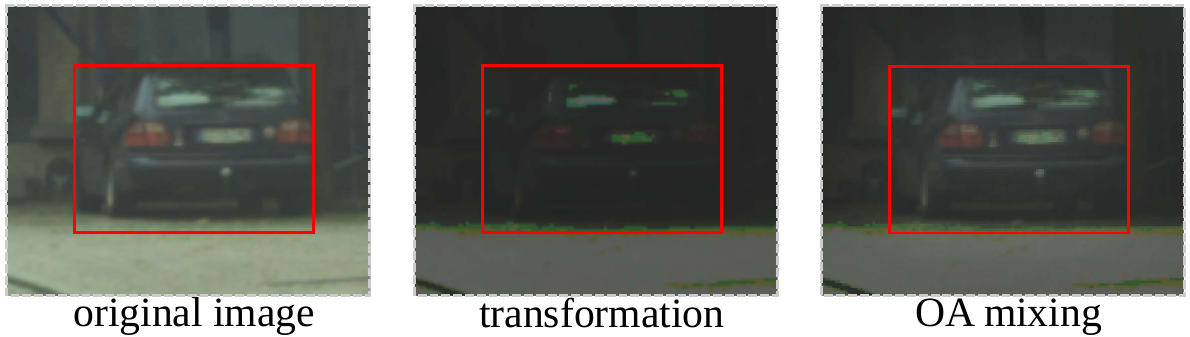}
        \caption{} 
        \label{fig:problem__data_augmentation b}
    \end{subfigure}
     \end{center}
     \caption{Overview of OA-Mix method. (a) Multi-level transformations generate locally diverse changes. Red and green boxes denote foreground and random regions, respectively. (b) Object-aware mixing strategy preserves semantic features against image transformations.
     } \label{fig:oamix}
\end{figure}

Diversity and affinity are two important factors to be considered in data augmentation for S-DGOD.
Diversity refers to the variety of features in the augmented image.
Affinity refers to the distributional similarity between the original and augmented images.
In other words, effective data augmentation methods should generate diverse data that does not deviate significantly from the distribution of the original data.
This section decomposes OA-Mix into two components: multi-level transformation for diversity and object-aware mixing strategy for affinity.

\paragraph{Multi-level transformations} \label{multilevel_transformations}
OA-Mix enhances domain diversity by applying locally diverse changes within an image. Locally diverse changes can be achieved with a multi-level transformation strategy. As shown in~\figref{fig:problem__data_augmentation a}, an image is randomly divided into several regions such as foreground, background, and random box regions.
Then, different transformation operations, such as color and spatial transformations, are randomly applied to each region.
Spatial transformation operations are applied at the foreground level to preserve the location of the object.
As a result, multi-level transformation enhances the domain diversity of the augmented image without damaging the object locations

\paragraph{Object-aware mixing} \label{objectaware_mixing} 
In object detection, each object in an image has different characteristics, such as size, location, and color distribution. Depending on these object-specific characteristics, transformations can damage the semantic features of an object.
For example, \figref{fig:problem__data_augmentation b} shows that image transformation can damage the semantic features of objects with monotonous color distribution.
Previous works mix the original and augmented images at the image level to mitigate the degradation of semantic features.
However, the method of mixing the entire image with the same weight does not sufficiently utilize object information.

% Therefore, we calculate the saliency score $s$ for each object based on the saliency map $\bold{M}\in\mathbb{R}^{H\times W}$ to consider the properties of each object. The saliency score is defined as:
Therefore, we calculate the saliency score $s$ for each object based on the saliency map to consider the properties of each object. The saliency score is calculated as $\bold{S}\in\mathbb{R}^{h\times w}$ from object with size $h\times w$:

\begin{equation} \label{eq:saliency_score}
    s = \frac{1}{hw}\sum_{x=1}^{w}{\sum_{y=1}^{h}{\bold{S}_{x,y}}}.
\end{equation}

Specifically, the saliency map is a spatial representation of the spectral residual, which is the unpredictable frequency in an image.
The object with a high saliency score has clear semantic signals.
In contrast, the object with a low score has weak semantic signals as shown in~\figref{fig:problem__data_augmentation b}.
Object-aware mixing strategy increases the mixing weight of the original image for objects with low saliency scores, thereby preventing semantic feature damage.
% \textcolor{blue}{Specifically, it linearly combines the original and augmented images with the mixing weight $m$:}
Specifically, for each area $P$ of the image, it linearly combines the original and augmented images with the mixing weight $m$:
% $$
% \bold{I}_\text{oamix} += \bold{M}\otimes\Big( m\bold{I}_\text{aug} + (1-m)\bold{I}_\text{orig} \Big),
% $$
% $$
% \bold{I}_\text{oamix}^P = m\bold{I}_\text{aug}^P + (1-m)\bold{I}_\text{orig}^P,
% $$
% $$
% \bold{I}_\text{oamix} = \sum_{P}\bold{I}_\text{oamix}^P
% $$
$$
\bold{I}_\text{oamix} = \sum_{P}{\Big(m\bold{I}_\text{aug}^P + (1-m)\bold{I}_\text{orig}^P}\Big),
$$
% \textcolor{blue}{
% \noindent where $\bold{I}_\text{oamix}$ is the result image, which is initialized with zeros. $\bold{M}$ is the mask for the divided region. $m$ is sampled from different distributions depending on saliency score $s$. This process is repeated until all areas of the result image are covered.}
\noindent where $m$ is sampled from different distributions depending on saliency score $s$.
As a result, the strategy enhances the affinity of the augmented image, mitigating the negative effects of transformations.

%-----------------------------------------------------------------------
\subsection{OA-Loss}
OA-Loss is designed to train the domain-invariant representations between the original and OA-Mixed domains in an object-aware approach. The object-aware method arranges instances in multi-domain according to the semantics of the instances. OA-Loss does not depend on any object detection framework and can be applied to both one-stage and two-stage detection frameworks.

\paragraph{Review of supervised contrastive learning}
The supervised contrastive learning methods construct positive pairs for the same class and negative pairs for different classes~\cite{khosla2020supervised, sun2021fsce, yao2022pcl}. The methods align intra-class and repulse inter-class instances in the embedding space. 
Previous methods for object detection~\cite{sun2021fsce} utilized only the foreground instances and ignored the semantic relations among background instances. In contrast, we explore the meaning of background instances for domain-invariant representations.

\paragraph{Meaning of background instances}
In object detection, each instance feature is labeled as the background if all intersection of unions (IoUs) with the ground-truth set are less than an IoU threshold, and foreground otherwise. Instance features correspond to region proposals and grid cell features in two-stage and one-stage detection frameworks, respectively.

\begin{comment}
\begin{figure}[t!]
    \begin{center}
        \includegraphics[width=0.44\textwidth]{AnonymousSubmission/Figure/bgiou/bgiou.png}
    \end{center}
    \caption{
    Ratio of background instances in Faster R-CNN model training. Bg denotes the background instances and accounted for $94\%$ of all the instances. For each background instance, the total IoU was calculated as the sum of the IoUs between the background instance and objects. The background instances with $\text{total IoU} \geq 0.3 $ accounted for $53\%$ of all the instances.}
    \label{fig:bgiou}
\end{figure}
\end{comment}

The background instances can partially contain the foreground object of different classes %as shown in~\figref{fig:bgiou}. 
However, existing supervised contrastive learning methods~\cite{khosla2020supervised, kang2021exploring, li2022targeted, yao2022pcl} regard different background instances identically and form positive pairs, leading to false semantic relations of foreground objects. Therefore, the relationship between background instances is required to be defined in an object-aware manner.

\paragraph{OA contrastive learning}
OA-Loss trains the semantic relations among instances in an object-aware approach to reduce the multi-domain gap.  For foreground instances, the method pulls the same class and pushes different classes away to improve object classification in multi-domain~\cite{sun2021fsce}. For background instances, the method pushes background instances away from each other except for the augmented instances as shown in \figref{fig:the overview of OA}. OA-Loss trains the model to discriminate between background instances containing foreground objects of various classes. As a result, our approach reflects the meaning of the background class, which is difficult to define as a single class. 

In addition, the negative pairs of background instances repulse each other, helping the model to output different features.
OA-Loss allows the model to generate various background features and trains the multi-domain gaps for various features, further improving generalization capability of the object detector.

To incorporate positive and negative sample pairs into contrastive learning, we encode the instance features into contrastive features~$\bold{Z}$. Following supervised contrastive learning~\cite{kim2021selfreg, sun2021fsce}, a contrastive branch is introduced parallel to the classification and regression branches as shown in \figref{fig:the overview of OA}. The contrastive branch consists of two layer multilayer perceptrons (MLPs). 
The contrastive branch generates feature set~$\bold{Z}$ from the multi-level regions of OA-Mix and proposed regions of the model.
The multi-level regions enable detectors to learn semantic relations in a wider variety of domains.

We set each contrastive feature~$\bold{z}_i\in \bold{Z}$ as an anchor. The positive set of each feature $\bold{z}_i$ is defined as:
{\footnotesize
\begin{equation}
    \bold{Z}^{pos}_i = 
    \begin{cases}
        \{ 
            \bold{z} | y(\bold{z}) = y(\bold{z}_i), \forall \bold{z}\in\bold{Z}\setminus \{ \bold{z}_i \}
        \} 
        &  \text{if foreground}  \\
        \{ \bold{z}'_i \} & \text{otherwise},\\
    \end{cases}
\end{equation}
}

\noindent where $y$ is the label mapping function for feature $\bold{z}$. $\bold{z}'$ is the augmented feature from $\bold{z}$. 
The positive set $\bold{Z}^{pos}$ is defined as the same class for foreground anchors and the same instance for background anchors.
The proposed contrastive loss~$L_{ct}$ is defined as:

%v0.2
\begin{equation}
    L_{ct}=\frac{1}
    {\left|\bold{Z}\right|}\sum_{\bold{z}_i\in\bold{Z}}{L_{\bold{z}_i}},
\end{equation}
\begin{equation}
    \small{
        L_{\bold{z}_i} 
        = \frac{-1}{|\bold{Z}^{pos}_i|}
        \sum_{\bold{z}_j\in\bold{Z}^{pos}_i}
        {\log{
            \frac{
            \exp{(\tilde{\bold{z}}_i\cdot\tilde{\bold{z}}_j/\tau)}}
            {\displaystyle\sum_{\bold{z}_k\in\bold{Z}\setminus\{\bold{z}_i\}}{\exp{(\tilde{\bold{z}}_i\cdot\tilde{\bold{z}}_k/\tau)}}}
                }},
    }
\end{equation}

\noindent where $|\bold{Z}|$ and $|\bold{Z}^{pos}_{i}|$ are the cardinalities of $\bold{Z}$ and $\bold{Z}^{pos}_{i}$, respectively. $\tau$ is the temperature scaling parameter and $\tilde{\bold{z}}_i=\frac{\bold{z}_i}{||\bold{z}_i||}$ denotes normalized features. The contrastive loss optimizes the feature similarity to align instances for the positive set $\bold{Z}^{pos}_{i}$, and repulse otherwise.

%------------------------------------------------------------------------ 

%%%%%%%%%%%%%%%%%%%%%%%%%%%%%%%%%%%%%%%%%%%
%%% TABLE 1. Cityscapes %%%
%%%%%%%%%%%%%%%%%%%%%%%%%%%%%%%%%%%%%%%%%%%

\begin{table*}[!tbp] 
    \begin{center}
    \begin{adjustbox}{max width=\textwidth}
    \begin{threeparttable}
        {\fontsize{28}{33.6}\selectfont
        %\Huge{% \footnotesize{
        % Column: [Method | Clean | Gauss.|Shot|Impulse | Defocus|Glass|Motion|Zoom | Snow|Frost|Fog|Bright | Contrast|Elastic|Pixel|JPEG' | mPC]
        \begin{tabular}
            {@{}r  *{1}{c}@{} 
                *{3}{c}@{} | *{4}{c}@{} | *{4}{c}@{} | *{4}{c}@{} | *{1}{c}@{} }
            % \T\B \\ \hline
            \\ \hline
            \toprule 
                & % & \multicolumn{1}{c@{}}{\multirow{2}{*}{Clean}}
                & \multicolumn{3}{c@{}}{Noise}
                & \multicolumn{4}{c@{}}{Blur}
                & \multicolumn{4}{c@{}}{Weather}
                & \multicolumn{4}{c@{}}{Digital}
            \\ % & \multicolumn{1}{c@{}}{\multirow{2}{*}{mPC}}\\ 
            \cmidrule(lr){3-5} \cmidrule(lr){6-9} \cmidrule(lr){10-13} \cmidrule(lr){14-17} %\cmidrule(lr){2-3} % from 1'st to 18'th column
                \multicolumn{1}{c|@{}}{Method} % \mc{Method} % \multicolumn{1}{c|@{}}{Method}
                & \multicolumn{1}{c|@{}}{Clean} % \mc{Clean} % \multicolumn{1}{c|@{}}{Clean}
                & \mc{Gauss.} & \mc{Shot} & \mc{Impulse}
                & \mc{Defocus} & \mc{Glass} & \mc{Motion} & \mc{Zoom}
                & \mc{Snow} & \mc{Frost} & \mc{Fog} & \mc{Bright}
                & \mc{Contrast} & \mc{Elastic} & \mc{Pixel} & \mc{JPEG'}
                & \multicolumn{1}{|c@{}}{mPC} % \mc{mPC} % \multicolumn{1}{|c@{}}{mPC}
            \\
            % \cmidrule(lr){1-1} \cmidrule(lr){2-2} \cmidrule(lr){3-17} \cmidrule(lr){18-18}
            \hline \noalign{\vskip 0.7mm}
                \rowcolor{gray!20}
                \multicolumn{1}{r|@{}}{Standard} % AI28/faster_rcnn_r50_fpn_1x_cityscapes
                & \multicolumn{1}{c|@{}}{42.2} % Clean
                & \mc{0.5} & \mc{1.1} & \mc{1.1}  % Noise
                & \mc{17.2} & \mc{16.5} & \mc{18.3} & \mc{2.1} % Blur
                & \mc{2.2} & \mc{12.3} & \mc{29.8} & \mc{32.0} % Weather
                & \mc{24.1} & \mc{40.1} & \mc{18.7} & \mc{15.1} % Digital
                & \multicolumn{1}{|c@{}}{15.4} % mPC
            \\ % \addlinespace
            \midrule
            \multicolumn{18}{l@{}}{\textbf{+ Data augmentation}} \\
            \midrule
            %     \multicolumn{1}{r|@{}}{Mosaic} % VPR1/other_augs/city_mosaic
            %     & \multicolumn{1}{c|@{}}{31.6} % Clean
            %     & \mc{0.4}  & \mc{0.9}  & \mc{0.2}  % Noise
            %     & \mc{12.0} & \mc{12.4} & \mc{12.2} & \mc{0.9} % Blur
            %     & \mc{1.4}  & \mc{8.6} & \mc{18.4} & \mc{20.5} % Weather
            %     & \mc{13.4} & \mc{29.1} & \mc{8.4} & \mc{7.9} % Digital
            %     & \multicolumn{1}{|c@{}}{9.8} % mPC      
            % \\ % \addlinespace
                \multicolumn{1}{r|@{}}{Cutout} % AI28/rebuttal_other_augs/city_cutout
                & \multicolumn{1}{c|@{}}{42.5} % Clean
                & \mc{0.6}  & \mc{1.2}  & \mc{1.2}  % Noise
                & \mc{17.8} & \mc{15.9} & \mc{18.9} & \mc{2.0} % Blur
                & \mc{2.5}  & \mc{13.6} & \mc{29.8} & \mc{32.3} % Weather
                & \mc{24.6} & \mc{40.1} & \mc{18.9} & \mc{15.6} % Digital
                & \multicolumn{1}{|c@{}}{15.7} % mPC      
            \\ % \addlinespace
                \multicolumn{1}{r|@{}}{Photo'} % VPR1/other_augs/city_photometricdistrotion
                & \multicolumn{1}{c|@{}}{\textbf{42.7}} % Clean
                & \mc{1.6}  & \mc{2.7}  & \mc{1.9}  % Noise
                & \mc{17.9} & \mc{14.1} & \mc{18.7} & \mc{2.0} % Blur
                & \mc{2.4}  & \mc{16.5} & \mc{36.0} & \mc{\textbf{39.1}} % Weather
                & \mc{27.1} & \mc{39.7} & \mc{18.0} & \mc{16.4} % Digital
                & \multicolumn{1}{|c@{}}{16.9} % mPC 
            \\ % \addlinespace
                \multicolumn{1}{r|@{}}{AutoAug-det~$\dagger$} % VPR1/other_augs/city_autoaugdet_seed7
                & \multicolumn{1}{c|@{}}{42.4} % Clean
                & \mc{0.9} & \mc{1.6} & \mc{0.9}  % Noise
                & \mc{16.8} & \mc{14.4} & \mc{18.9} & \mc{2.0} % Blur
                & \mc{1.9} & \mc{16.0} & \mc{32.9} & \mc{35.2} % Weather
                & \mc{26.3} & \mc{39.4} & \mc{17.9} & \mc{11.6} % Digital
                & \multicolumn{1}{|c@{}}{15.8} % mPC
            % \\ % \addlinespace
            %     \multicolumn{1}{r|@{}}{Stylized} % AI28/augment2/city_sin_none_rpn.none.none_roi.none.none__e2
            %     & \multicolumn{1}{c|@{}}{33.4} % Clean
            %     & \mc{9.0} & \mc{11.2} & \mc{8.7}  % Noise
            %     & \mc{20.9} & \mc{19.1} & \mc{17.6} & \mc{4.2} % Blur
            %     & \mc{4.7} & \mc{17.6} & \mc{28.0} & \mc{30.5} % Weather
            %     & \mc{21.9} & \mc{31.6} & \mc{26.8} & \mc{22.6} % Digital
            %     & \multicolumn{1}{|c@{}}{18.3} % mPC     
            \\ % \addlinespace
                \multicolumn{1}{r|@{}}{AugMix} % AI28/augment/city_augmix.augs_none_rpn.none.none_roi.none.none__e2
                & \multicolumn{1}{c|@{}}{39.5} % Clean
                & \mc{5.0}  & \mc{6.8}  & \mc{5.1}  % Noise
                & \mc{18.3} & \mc{18.1} & \mc{19.3} & \mc{\textbf{6.2}} % Blur
                & \mc{5.0}  & \mc{20.5} & \mc{31.2} & \mc{33.7} % Weather
                & \mc{25.6} & \mc{37.4} & \mc{20.3} & \mc{19.6} % Digital
                & \multicolumn{1}{|c@{}}{18.1} % mPC      
            \\ % \addlinespace
                \multicolumn{1}{r|@{}}{Stylized}
                & \multicolumn{1}{c|@{}}{36.3} % Clean
                & \mc{4.8} & \mc{6.8} & \mc{4.3}  % Noise
                & \mc{19.5} & \mc{18.7} & \mc{18.5} & \mc{2.7} % Blur
                & \mc{3.5} & \mc{17.0} & \mc{30.5} & \mc{31.9} % Weather
                & \mc{22.7} & \mc{33.9} & \mc{\textbf{22.6}} & \mc{\textbf{20.8}} % Digital
                & \multicolumn{1}{|c@{}}{17.2} % mPC
            \\ % \addlinespace
                \rowcolor{gray!20}
                \multicolumn{1}{r|@{}}{OA-Mix (ours)} % Hendrycks/oa_augmix5/city_pickoam_saliency_sparse_augmix_s10
                & \multicolumn{1}{c|@{}}{\textbf{42.7}} % Clean
                & \mc{\textbf{7.2}} & \mc{\textbf{9.6}} & \mc{\textbf{7.7}}  % Noise
                & \mc{\textbf{22.8}} & \mc{\textbf{18.8}} & \mc{\textbf{21.9}} & \mc{5.4} % Blur
                & \mc{\textbf{5.2}} & \mc{\textbf{23.6}} & \mc{\textbf{37.3}} & \mc{38.7} % Weather
                & \mc{\textbf{31.9}} & \mc{\textbf{40.2}} & \mc{22.2} & \mc{20.2} % Digital
                & \multicolumn{1}{|c@{}}{\textbf{20.8}} % mPC
            \\ % \addlinespace
            \midrule
            \multicolumn{18}{l@{}}{\textbf{+ Loss function}} \\
            \midrule
                \multicolumn{1}{r|@{}}{SupCon}
                & \multicolumn{1}{c|@{}}{43.2} % Clean
                & \mc{7.0} & \mc{9.5} & \mc{7.4}  % Noise
                & \mc{22.6} & \mc{20.2} & \mc{22.3} & \mc{4.3} % Blur
                & \mc{5.3} & \mc{23.0} & \mc{37.3} & \mc{38.9} % Weather
                & \mc{31.6} & \mc{40.1} & \mc{\textbf{24.0}} & \mc{20.1} % Digital
                & \multicolumn{1}{|c@{}}{20.9} % mPC
            \\ % \addlinespace
                \multicolumn{1}{r|@{}}{FSCE}
                & \multicolumn{1}{c|@{}}{43.1} % Clean
                & \mc{7.4} & \mc{10.2} & \mc{8.2}  % Noise
                & \mc{23.3} & \mc{20.3} & \mc{21.5} & \mc{4.8} % Blur
                & \mc{5.6} & \mc{23.6} & \mc{37.1} & \mc{38.0} % Weather
                & \mc{31.9} & \mc{40.0} & \mc{23.2} & \mc{20.4} % Digital
                & \multicolumn{1}{|c@{}}{21.0} % mPC
            \\ % \addlinespace
                \rowcolor{gray!20}
                \multicolumn{1}{r|@{}}{\makecell[r]{OA-Loss (ours) \\ = OA-DG}}
                & \multicolumn{1}{c|@{}}{\textbf{43.4}} % Clean
                & \mc{\textbf{8.2}} & \mc{\textbf{10.6}} & \mc{\textbf{8.4}}  % Noise
                & \mc{\textbf{24.6}} & \mc{\textbf{20.5}} & \mc{\textbf{22.3}} & \mc{4.8} % Blur
                & \mc{\textbf{6.1}} & \mc{\textbf{25.0}} & \mc{\textbf{38.4}} & \mc{\textbf{39.7}} % Weather
                & \mc{\textbf{32.8}} & \mc{\textbf{40.2}} & \mc{23.8} & \mc{\textbf{22.0}} % Digital
                & \multicolumn{1}{|c@{}}{\textbf{21.8}} % mPC
            \\ % \addlinespace
            \bottomrule
        \end{tabular}
        }
        \begin{tablenotes} 
        {\fontsize{28}{33.6}\selectfont
        % \Huge% \small
        \item[$\dagger$] We followed the searched policies~\cite{zoph2020learning}.}
        % We followed the searched policies on COCO dataset~\cite{zoph2020learning}.
        \end{tablenotes}
    \end{threeparttable}
    \end{adjustbox}
    \end{center}
    \caption{Comparison with state-of-the-art methods on Cityscapes-C. For each corruption type, the average performance was calculated. mPC is an average performance of 15 corruption types.} \label{table1}
    \end{table*} 

We also designed the consistency loss to reduce the gap between the original and augmented domains at logit-level. To make consistent output in multi-domain~\cite{hendrycksaugmix, modas2022prime}, we adopt Jensen-Shannon (JS) divergence as the consistency loss to reduce the gap between the original and OA-Mixed domains. JS divergence is a symmetric and smoothed version of the Kullback–Leibler (KL) divergence. The consistency loss is defined as:

\begin{equation} 
    L_{cs}
    = \frac{1}{2}(\textbf{KL}[\mathbf{p} || \mathbf{M}] + \textbf{KL}[\mathbf{p}_{+} || \mathbf{M}]),
\end{equation}
\noindent where $\mathbf{p}$ and $\mathbf{p}_{+}$ are model predictions from original and OA-Mixed images, respectively. $\mathbf{M} = \frac{1}{2} (\mathbf{p}+\mathbf{p}_{+})$ is the mixture probability of predictions. The loss improves the generalization ability to classify the objects in OOD.

%------------------------------------------------------------------------

\subsubsection{Training objectives}
Our method trains the base detector in an end-to-end manner. Our OA-Loss~$L_{OA}$ consists of consistency loss~$L_{cs}$ and contrastive loss~$L_{ct}$. 
\begin{equation}
    L_{OA} = L_{cs} + \gamma L_{ct},
\end{equation}
\noindent where $\gamma$ is hyperparameter. OA-Loss can be added to the original loss~$L_{det}$ for the general detector. The joint training loss is defined as
\begin{equation}
    L = L_{det} + \lambda L_{OA},
\end{equation}
\noindent where $\lambda$ balances the scales of the losses. The joint training loss allows the model to learn object semantics and domain-invariant representations from OA-Mixed domains.

\section{Experiments}   
In this section, we evaluate the robustness of our method against out-of-distribution. We also conduct ablation studies to verify the effectiveness of proposed modules. 
%-------------------------------------------------------------------------
\subsection{Datasets}
We evaluated the DG performance of our method in an urban scene for common corruptions and various weather conditions. Cityscapes-C~\cite{michaelis2019benchmarking} is a test benchmark to evaluate object detection robustness to corrupted domains. Cityscapes-C provides 15 corruptions on five severity levels. The corruptions are divided into four categories: noise, blur, weather, and digital. These corruption types are used to measure and understand the robustness against OOD~\cite{hendrycks2018benchmarking}. The common corruptions should be used only to evaluate the robustness of the model and are strictly prohibited to be used during training.

Diverse Weather Dataset (DWD) is an urban-scene detection benchmark to assess object detection robustness to various weather conditions. DWD collected data from BDD-100k~\shortcite{yu2020bdd100k}, FoggyCityscapes~\shortcite{sakaridis2018semantic}, and Adverse-Weather~\shortcite{hassaballah2020vehicle} datasets. It consists of five different weather conditions: daytime-sunny, night-sunny, dusk-rainy, night-rainy, and daytime-foggy. Training should be conducted only using the daytime-sunny dataset and robustness is evaluated against other adverse weather datasets.

\subsection{Evaluation metrics}
Following~\cite{michaelis2019benchmarking, wu2022single}, we evaluate the domain generalization performance of our method in various domains using mean average precision (mAP). The robustness against OOD is evaluated using mean performance under corruption (mPC), which is the average of mAPs for all corrupted domains and severity levels.
\begin{equation}
    \text{mPC} = \frac{1}{N_C}\sum_{C=1}^{N_C}{        
            \frac{1}{N_S}\sum_{S=1}^{N_S}{P_{C,S}}
        },
\end{equation}
\noindent where $P_{C,S}$ is the performance on the test data corrupted by corruption~$C$ at severity level~$S$, $N_C$ and $N_S$ are the number of corruption and severity, respectively. $N_C$ and $N_S$ are set to 15 and 5 in Cityscapes-C; and 4 and 1 in DWD, respectively.

\subsection{Robustness on common corruptions}
\tabref{table1} shows the performance of the state-of-the-art models on clean and corrupted domains. The baseline model is Faster R-CNN with ResNet-50 and feature pyramid networks (FPN). Each model is trained only on the clean domain and evaluated on both the clean and corrupted domains. Temperature scaling parameter $\tau$ for contrastive loss is set to $0.06$. We set $\lambda$ and $\gamma$ to $10$ and $0.001$. More details can be seen in the supplementary material.

Cutout~\cite{zhong2020random} and photometric distortion~\cite{redmon2018yolov3} are data augmentation methods that can improve the generalization performance of object detectors. However, they showed limited improvements in performance on corrupted domains.
AutoAug-det~\cite{zoph2020learning} explores augmentation policies for the generalization of object detection.
The method improved performance in the clean domain, but there was no significant performance improvement in the corrupted domain.

AugMix~\cite{hendrycksaugmix} and stylized augmentation~\cite{geirhosimagenet} are effective for S-DG in image classification. AugMix generates augmented images with unclear object locations, leading to a performance drop on the clean domain. Stylized augmentation improved performance on the corrupted domains with style transfer. However, the method did not consider affinity with the original domain, which decreased performance in the clean domain.

In contrast, OA-Mix maintains the clean domain performance with an object-aware mixing and improves the DG performance with multi-level transformations. 
Furthermore, we evaluated contrastive methods combined with OA-Mix. SupCon~\cite{khosla2020supervised} and FSCE~\cite{sun2021fsce} are contrastive learning methods that can reduce the multi-domain gap. The methods improved the clean performance, but the performance for the corrupted domain was not improved significantly. OA-Loss trains the semantic relations in an object-aware approach and achieved higher performance in both clean and corrupted domains. Consequently, OA-DG showed the best performance with $43.4$ mAP and $21.8$ mPC in the clean and corrupted domains, respectively.

%%%%%%%%%%%%%%%%%%%%%%%
%%% TABLE 2. S-DGOD %%%
%%%%%%%%%%%%%%%%%%%%%%%
\begin{table}[!tbp] 
    \begin{center}
    \begin{adjustbox}{max width=0.48\textwidth}
    \begin{threeparttable}
        {\fontsize{13.23}{15.876}\selectfont
        %\footnotesize{
        %\Large{
            \begin{tabular}{@{} r *{6}{c}@{}}
                    \toprule 
                        & \multicolumn{5}{|c|@{}}{mAP}
                        & \multicolumn{1}{c@{}}{\multirow{3}{*}{mPC}}\\ 
                    \cmidrule(lr){2-6}
                        & \multicolumn{1}{|c|@{}}{Daytime-} & \mc{Night-} & \mc{Dusk-} & \mc{Night-} & \multicolumn{1}{c|@{}}{Daytime-}\\
                        & \multicolumn{1}{|c|@{}}{sunny} & \mc{sunny} & \mc{rainy} & \mc{rainy} & \multicolumn{1}{c|@{}}{foggy}\\
                    \midrule
                    \multicolumn{1}{r|@{}}{Baseline} & \multicolumn{1}{c|@{}}{48.1} & 34.4 & 26.0 & 12.4 & 32.0 & \multicolumn{1}{|c@{}}{26.2}\\
                    \multicolumn{1}{r|@{}}{SW} & \multicolumn{1}{c|@{}}{50.6} & 33.4 & 26.3 & 13.7 & 30.8 & \multicolumn{1}{|c@{}}{26.1}\\
                    \multicolumn{1}{r|@{}}{IBN-Net} & \multicolumn{1}{c|@{}}{49.7} & 32.1 & 26.1 & 14.3 & 29.6 & \multicolumn{1}{|c@{}}{25.5} \\
                    \multicolumn{1}{r|@{}}{IterNorm} & \multicolumn{1}{c|@{}}{43.9} & 29.6 & 22.8 & 12.6 & 28.4 & \multicolumn{1}{|c@{}}{23.4}\\
                    \multicolumn{1}{r|@{}}{ISW} & \multicolumn{1}{c|@{}}{51.3} & 33.2 & 25.9 & 14.1 & 31.8 & \multicolumn{1}{|c@{}}{26.3}\\
                    \multicolumn{1}{r|@{}}{SHADE} & \multicolumn{1}{c|@{}}{-} & 33.9 & 29.5 & 16.8 & 33.4 & \multicolumn{1}{|c@{}}{28.4}\\
                    \multicolumn{1}{r|@{}}{CDSD} & \multicolumn{1}{c|@{}}{\textbf{56.1}} & 36.6 & 28.2 & 16.6 & 33.5 & \multicolumn{1}{|c@{}}{28.7}\\
                    \multicolumn{1}{r|@{}}{SRCD} & \multicolumn{1}{c|@{}}{-} & 36.7 & 28.8 & \textbf{17.0} & 35.9 & \multicolumn{1}{|c@{}}{29.6}\\
                    \midrule
                    \multicolumn{1}{r|@{}}{Ours} & \multicolumn{1}{c|@{}}{55.8} & \textbf{38.0} & \textbf{33.9} & 16.8 & \textbf{38.3} & \multicolumn{1}{|c@{}}{\textbf{31.8}}\\ % \addlinespace
                    \bottomrule 
            \end{tabular}%
        }
    \end{threeparttable}
    \end{adjustbox}
    \end{center}
    \caption{Comparison with state-of-the-art methods on Diverse Weather Dataset. mPC is an average performance of OOD weathers: night-sunny, dusk-rainy, night-sunny, and daytime-foggy.} \label{table2}
\end{table}

\subsection{Robustness on various weather conditions}
\tabref{table2} shows the DG performances in real-world weather conditions through DWD. We follow the settings in the CDSD~\cite{wu2022single} for DWD. We used Faster R-CNN with ResNet-101 backbone as a base object detector. Temperature scaling hyperparameter $\tau$ is set to $0.07$. We set $\lambda$ and $\gamma$ to $10$ and $0.001$, respectively, for Faster R-CNN. More details can be seen in the supplementary material.

OA-DG is compared with state-of-the-art S-DGOD methods. SW~\cite{pan2019switchable}, IBN-Net~\cite{pan2018two}, IterNorm~\cite{huang2019iterative}, ISW~\cite{choi2021robustnet}, and SHADE~\cite{zhao2022style} are feature normalization methods to improve DG. Compared with the baseline without any S-DGOD approaches, the feature normalization methods improve performance in certain weather conditions, but do not consistently enhance the performance for OOD.  
CDSD~\cite{wu2022single} and SRCD~\cite{rao2023srcd} improved the performances in all weather conditions with domain-invariant features. Our proposed OA-DG achieved the top performance with 31.8 mPC in OOD weather conditions. Compared with the baseline, our method improves domain generalization in all weather conditions, especially in corrupted environments such as dusk-rainy and daytime-foggy.

\subsection{Ablation studies}
All ablation studies evaluate DG performance in Cityscapes-C with the same settings as~\tabref{table1}.
Details and more ablation studies can be found in the supplementary material.

\begin{figure*}[t!]
    \begin{center}
        \makebox[0pt]{\rotatebox{90}{\hspace{1em}Baseline}\hspace*{2em}}%
        \begin{subfigure}{0.23\textwidth}
            \centering\includegraphics[width=\textwidth]{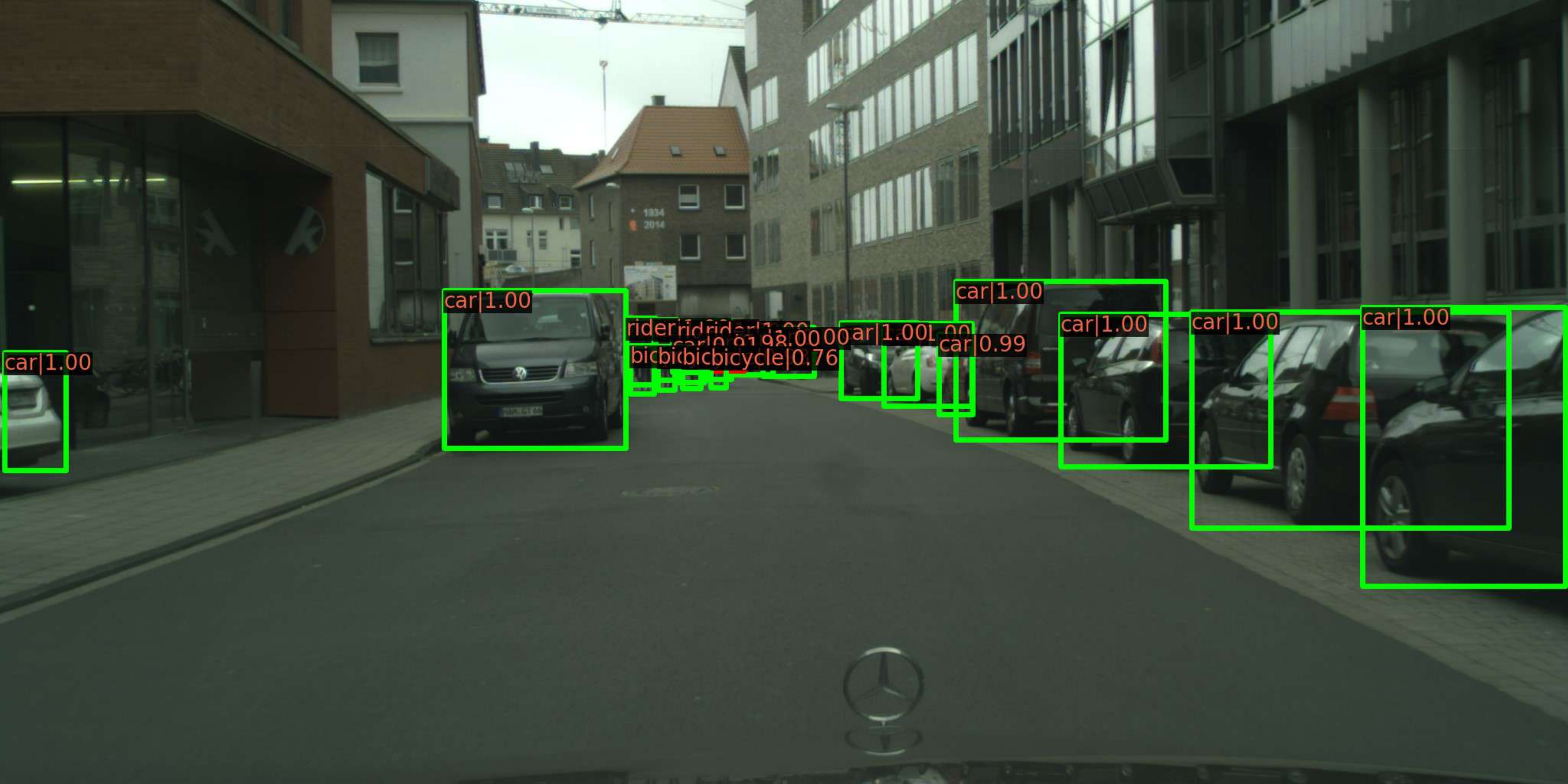}
        \end{subfigure}
        \begin{subfigure}{0.23\textwidth}
            \centering\includegraphics[width=\textwidth]{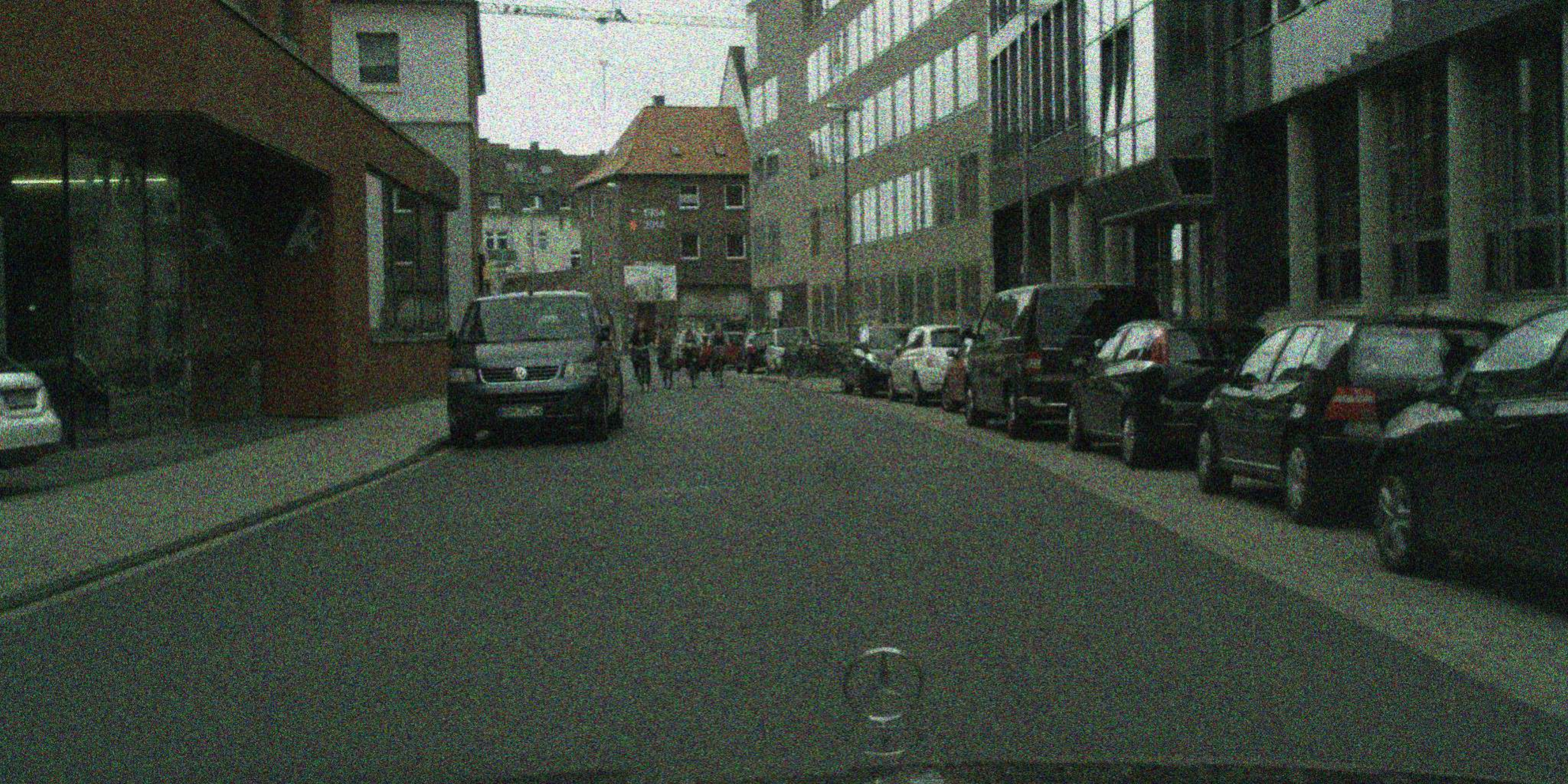}
        \end{subfigure}
        \begin{subfigure}{0.23\textwidth}
            \centering\includegraphics[width=\textwidth]{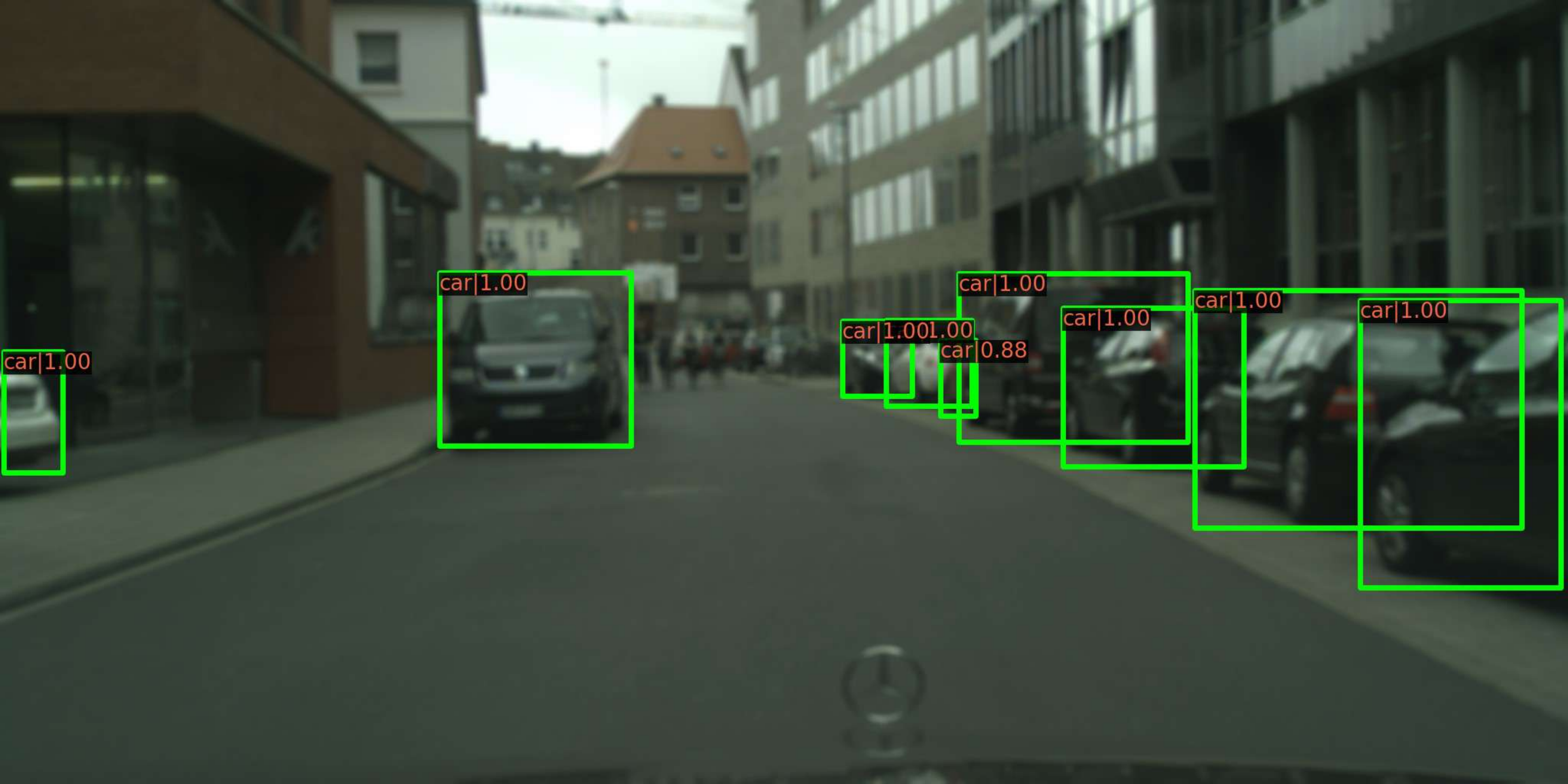}
        \end{subfigure}
        \begin{subfigure}{0.23\textwidth}
            \centering\includegraphics[width=\textwidth]{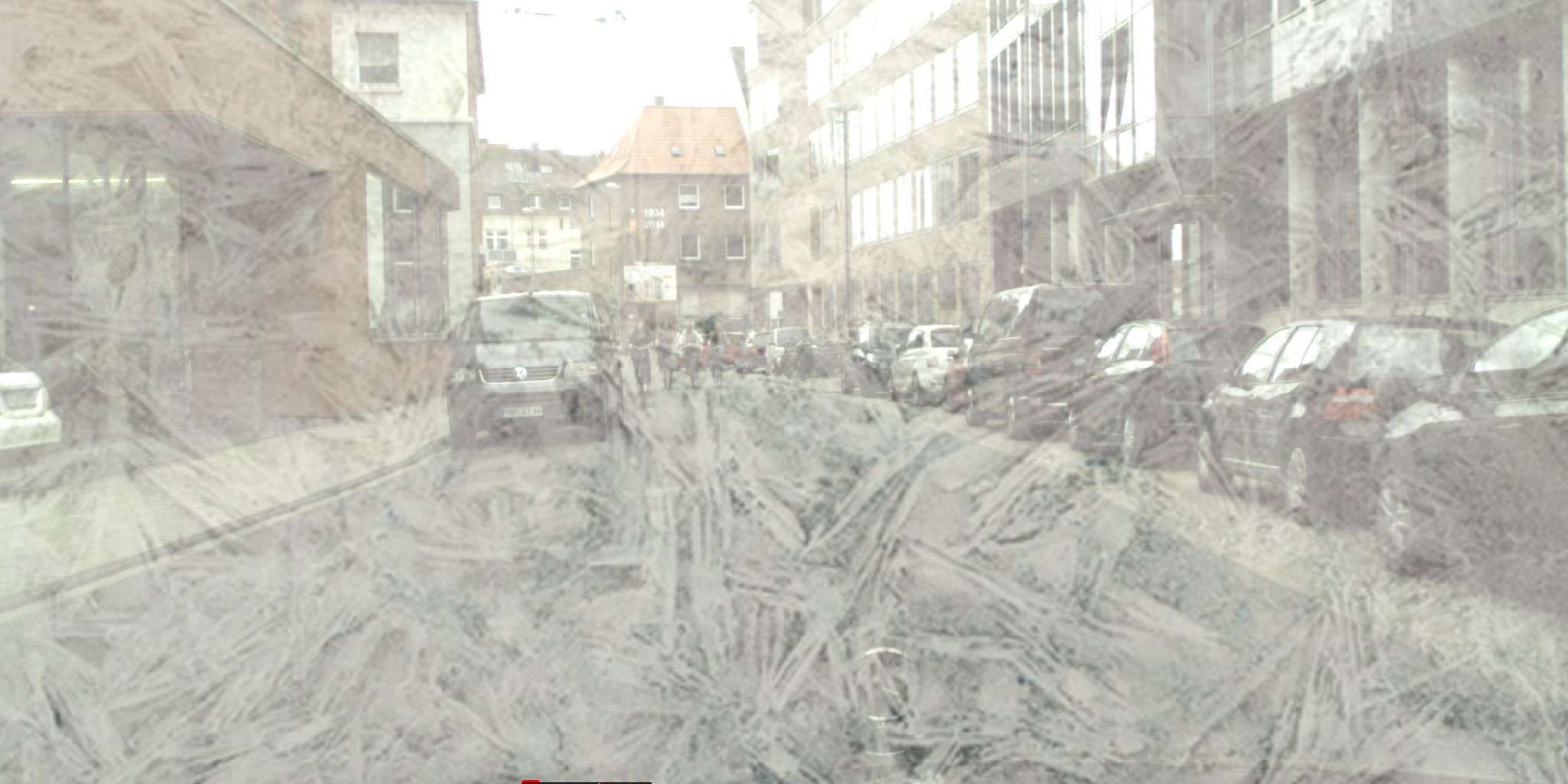}
        \end{subfigure}
        
        \makebox[0pt]{\rotatebox{90}{OA-DG (Ours)}\hspace*{2em}}%
        \begin{subfigure}{0.23\textwidth}
            \centering\includegraphics[width=\textwidth]{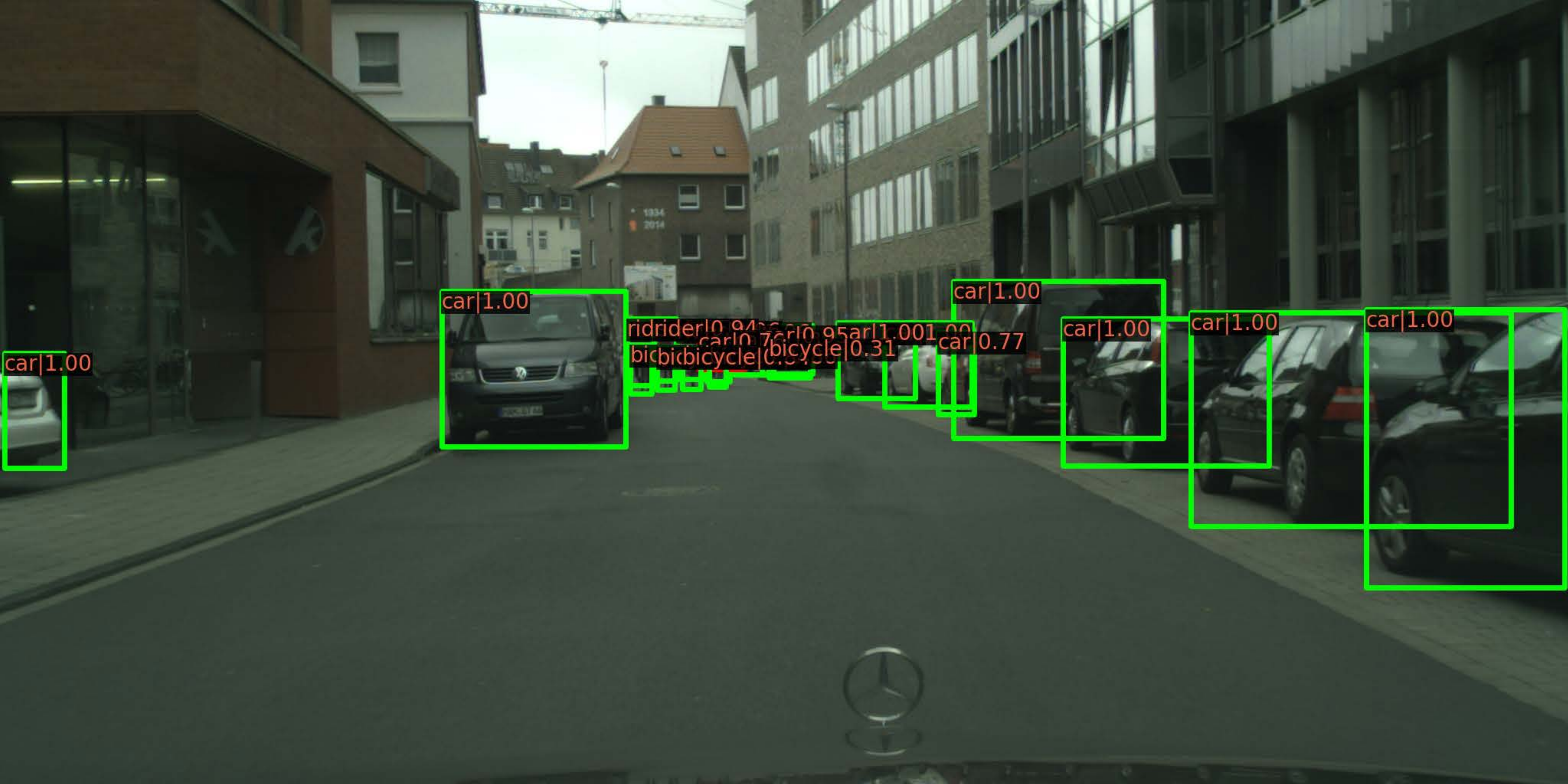}
        \end{subfigure}
        \begin{subfigure}{0.23\textwidth}
            \centering\includegraphics[width=\textwidth]{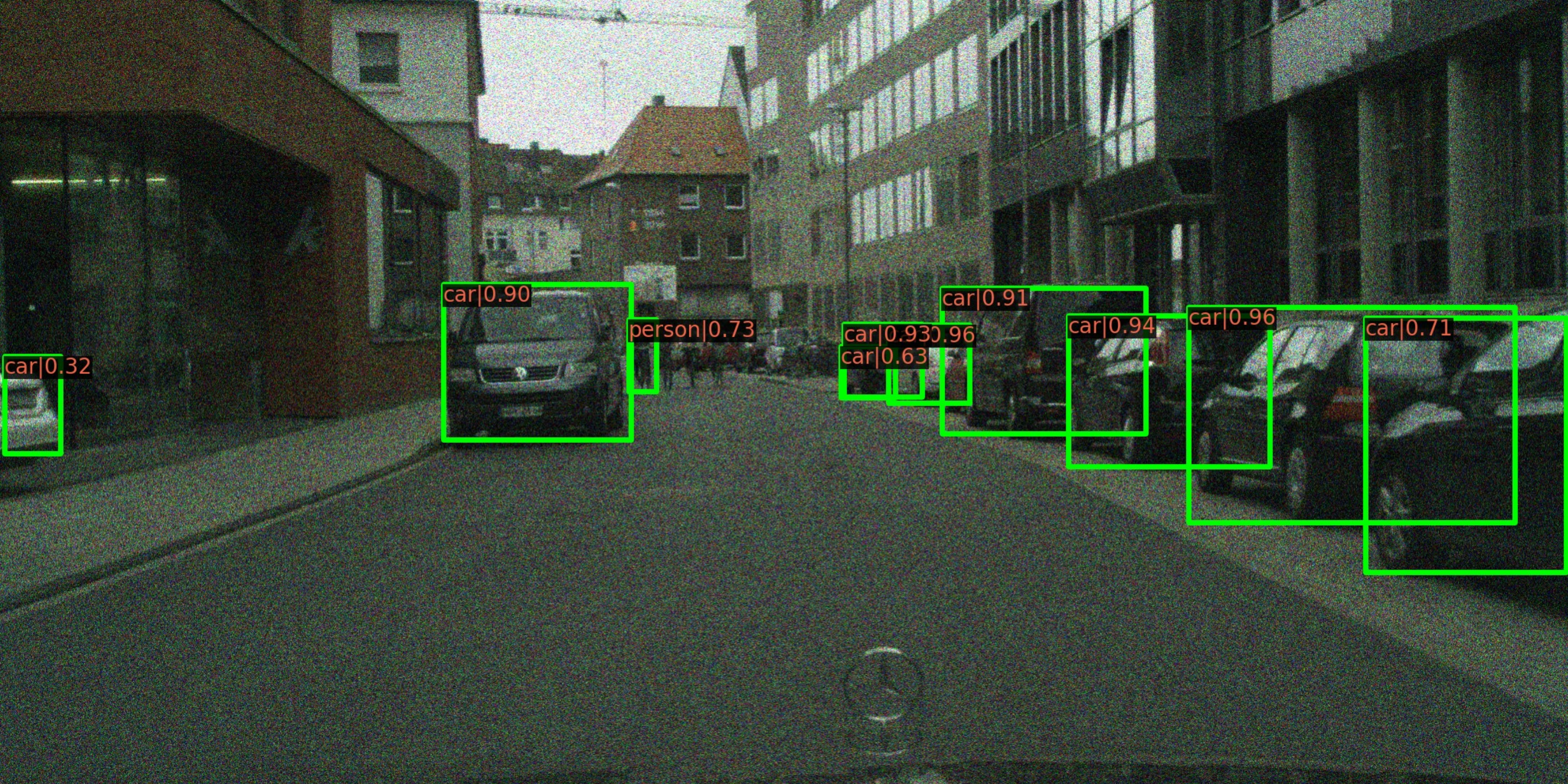}
        \end{subfigure}
        \begin{subfigure}{0.23\textwidth}
            \centering\includegraphics[width=\textwidth]{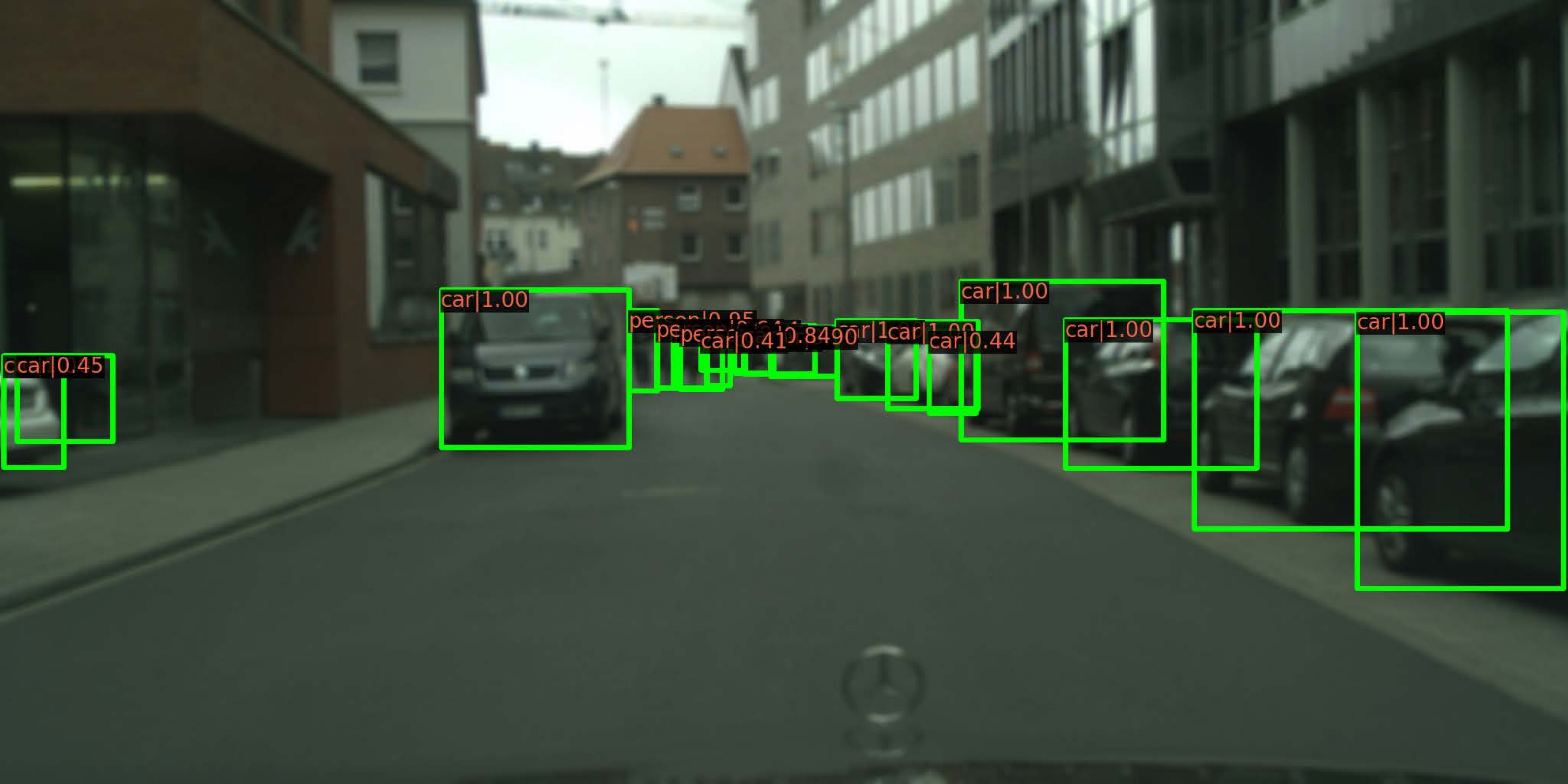}
        \end{subfigure}
        \begin{subfigure}{0.23\textwidth}
            \centering\includegraphics[width=\textwidth]{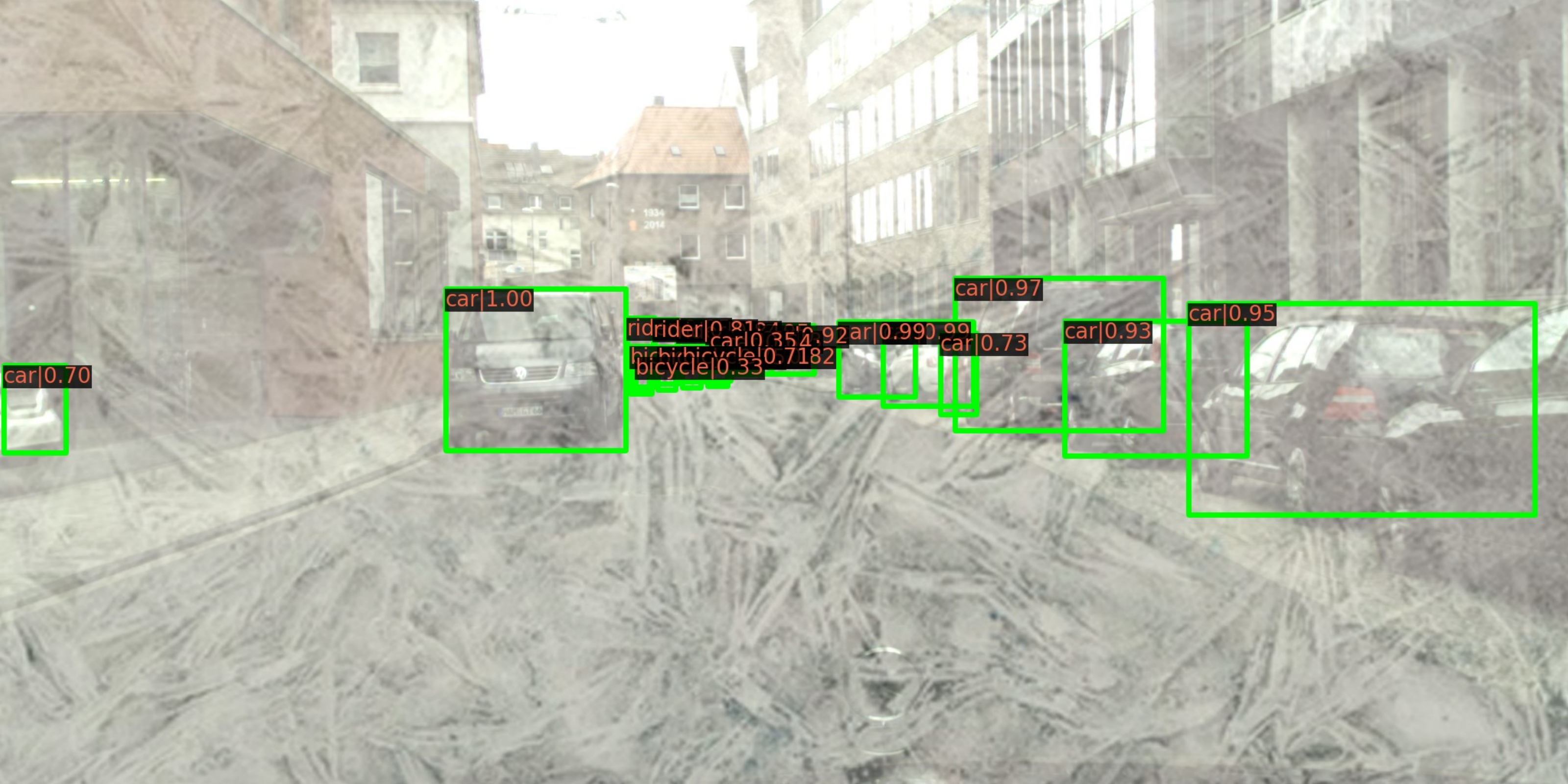}
        \end{subfigure}
    \end{center}
    \caption{From the left to the right, the detection results of the clean, shot noise, defocus blur, and frost domains are visualized.  
    Compared with the baseline, OA-DG detects the object more accurately in diverse corrupted domains.} \label{fig:result}
\end{figure*}
%%%%%%%%%%%%%%%%%%%%%%%%%%%%%%%%%%
%%% TABLE 3. Data augmentation %%%
%%%%%%%%%%%%%%%%%%%%%%%%%%%%%%%%%%

\begin{table}[!tbp] 
\begin{center}
    %{\fontsize{5}{6}\selectfont %\small
        % \setlength\tabcolsep{4pt}
        {\small
        \begin{tabular}{@{} c *{3}{c}@{}}
            \toprule 
                Transformations & Mixing strategy   & mAP   & mPC\\
            \midrule
                \xmark          & \xmark            & 42.2  & 15.4 \\
                Single-level    & \xmark            & 39.9  & 16.8 \\
                Multi-level     & \xmark            & 40.5  & 17.5  \\
                Multi-level     & Standard          & 41.6  & 19.7 \\
            \midrule % \rowcolor{gray!20}
                Multi-level     & Object-aware      & \textbf{42.7} & \textbf{20.8} \\
            \bottomrule 
        \end{tabular}
    }
\end{center}
\caption{Ablation analysis of our proposed OA-Mix on Cityscapes-C dataset.} \label{table3}
\end{table}

%%%%%%%%%%%%%%%%%%%%%%%%%%%%%%%%%%
%%% TABLE 4. OA Loss %%%
%%%%%%%%%%%%%%%%%%%%%%%%%%%%%%%%%%
\begin{table}[!t] 
\begin{center}
    {\small %\footnotesize 
        \begin{tabular}{c  cc  cc}
            \toprule 
                \multicolumn{1}{c}{\multirow{2}{*}{Consistency}}
                & \multicolumn{2}{c}{Contrastive}
                & \multicolumn{1}{c}{\multirow{2}{*}{mAP}}
                & \multicolumn{1}{c}{\multirow{2}{*}{mPC}} \\
            \cmidrule(lr){2-3}
                 & Target & Rule &  &  \\
            \midrule
                \xmark      & \xmark &  \xmark  & 42.7  & 20.8 \\
                \checkmark  & \xmark  & \xmark  & 43.1  & 21.2 \\
                \checkmark  & fg    & class-aware  &  \textbf{43.4} & 21.3 \\ 
                \checkmark  & fg+bg & class-aware  & 43.3  & 21.1 \\
                \checkmark  & fg+bg & object-aware & \textbf{43.4}  & \textbf{21.8} \\
            \bottomrule
        \end{tabular}
    }
\end{center}
\caption{Ablation analysis of our proposed contrastive method on Cityscapes-C. Our method considers all the semantic relations of the same instance, foreground class, background class, and background instances. fg and bg denote foreground and background instances, respectively.} 
\label{table4}
\end{table}

\subsubsection{OA-Mix}
This experiment validates the impact of individual components within OA-Mix. 
\tabref{table3} shows the clean performance and robustness according to multi-level transformations and object-aware mixing strategy.
The single-level transformation improved performance on corrupted domains compared with the baseline.
The multi-level transformation transforms the image at the local level and improved mPC compared with the single-level transformation.
However, both transformations showed lower clean performance than the baseline due to the damage to the semantic features of objects.
Standard mixing mitigates the negative effects of transformations. However, it does not utilize object information, which leads to lower clean performance than the baseline.
Only the object-aware mixing strategy showed a clean performance above the baseline and achieved the best mPC performance as well.

\begin{table}[!t]
   \begin{center}
      \begin{threeparttable}
          \small{
             \begin{tabular}[h]{r cc cc}
             \toprule
             & \multicolumn{2}{c@{}}{Faster R-CNN}
             & \multicolumn{2}{c@{}}{YOLOv3}\\
             \cmidrule(ll){2-3} \cmidrule(ll){4-5}
             & Baseline & OA-DG & Baseline & OA-DG\\ 
             \midrule
             mAP & 42.2 & \textbf{43.4} (+1.2) & 34.6 & \textbf{37.2} (+2.6) \\
             mPC & 15.4 & \textbf{21.8} (+6.4) & 12.4 & \textbf{15.6} (+3.2) \\
             FLOPs (G) & 408.7 & 409.0 & 194.0 & 199.5\\
             \bottomrule
             \end{tabular}
          }
      \end{threeparttable}
   \end{center}
   \caption{Performance and computational complexity of Faster R-CNN and YOLOv3 on Cityscapes.} \label{table5}
\end{table}

\subsubsection{Contrastive methods} 
In \tabref{table4}, we verify the effectiveness of OA-Loss on Cityscapes-C. The consistency loss improved mPC by reducing the domain gap between the original and augmented instances. Then, we conducted an ablation study of contrastive learning according to the target and rule. The class-aware methods simply pull intra-class and push inter-class away. Both class-aware methods improved clean performance, but did not improve performance on the corrupted domains. In contrast, OA-Loss allows the model to train multi-domain gaps in an object-aware manner for all the instances. Our OA-Loss was improved by 0.7 mAP and 1.0 mPC in both clean and corrupted domains, respectively, validating the superiority of the object-aware approach.

\subsubsection{Object detector architectures}
\tabref{table5} shows the generalization capability of OA-DG to object detection framework.
We conducted additional experiments with YOLOv3, a popular one-stage object detector.
Although YOLOv3 used various bag of tricks such as photometric distortion to improve its generalization capability, OA-DG improved its performance in both clean and corrupted domains.
This implies that our method is not limited to a specific detector architecture and can be applied to general object detectors.

\subsection{Qualitative analysis}
\subsubsection{Visual analysis}
\figref{fig:result} shows the results for the baseline and our model at clean and corrupted domains on Cityscapes.
In the clean domain, both the baseline and our model accurately detect objects, but the baseline fails to detect most objects in the corrupted domain. Compared with the baseline, our model accurately detects small objects and improves object localization in OOD.

%\begin{comment} %%% THIS IS IMPORTANT PART. DON'T REMOVE!
\subsubsection{Feature analysis}
\begin{figure}[t!]
    \begin{center}
     \includegraphics[width=0.46\textwidth]{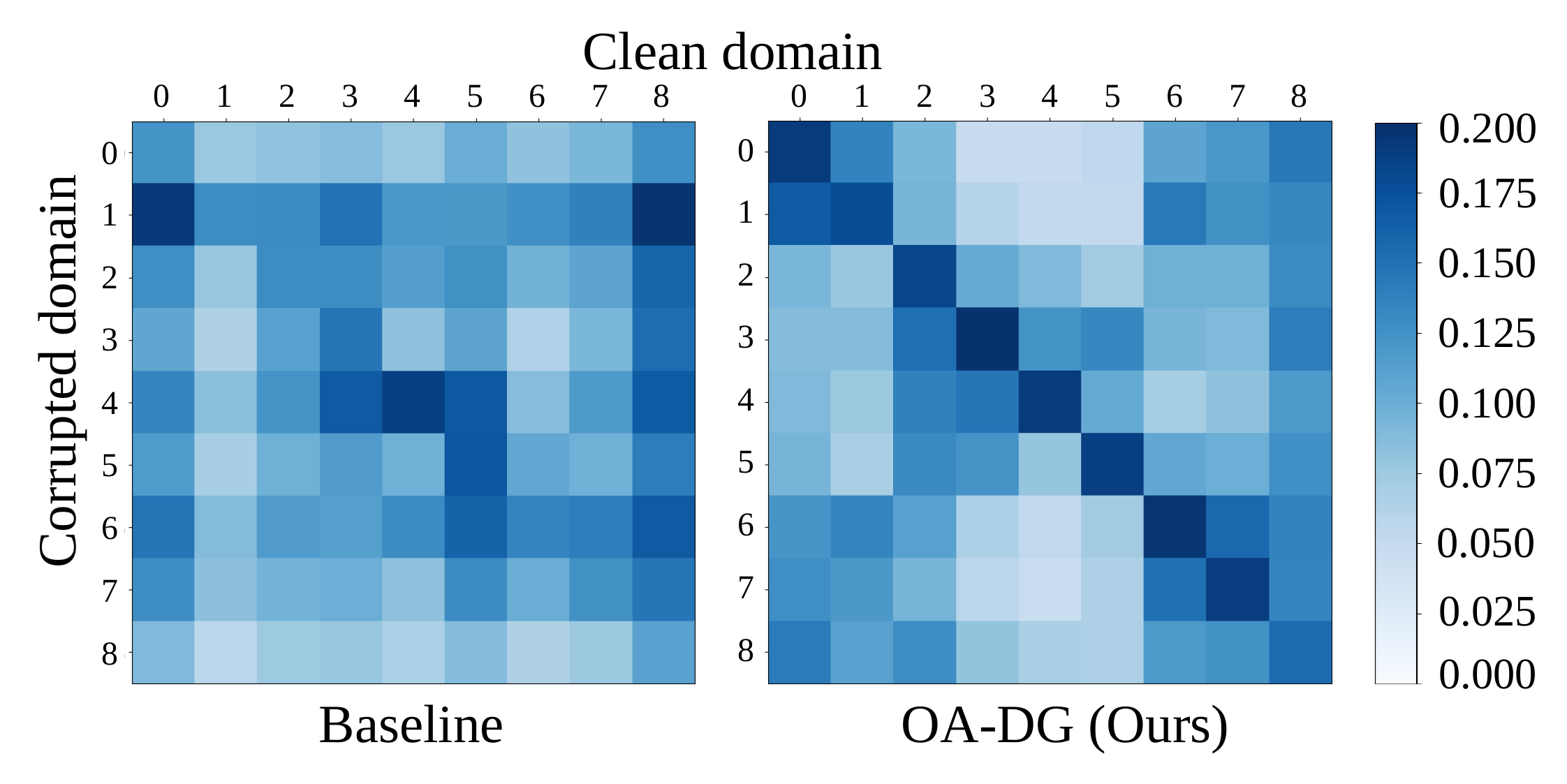}
    \end{center}  
    \caption{Feature correlations of the baseline and OA-DG
    between the clean and corrupted domains. Zero to seven on each axis denote the object classes and eight denotes the background class. 
    Compared with the baseline, our method has higher correlations between the same class in OOD.} \label{fig:feature correlations}
\end{figure}

\figref{fig:feature correlations} shows the feature correlations between the clean and shot noise domains on Cityscapes. The feature correlations are measured as the average of cosine similarity for each class and normalized in the $x$-axis. Following~\cite{ranasinghe2021orthogonal}, we use the features of the penultimate layer in the classification head. 
In \figref{fig:feature correlations}, the $x$-axis and $y$-axis represent the clean and corrupted domains, respectively.
The comparison between the corrupted and clean domains shows that the baseline has little correlation between the same classes. This hinders the baseline from detecting objects in the corrupted domain, as shown in \figref{fig:result}. Compared with the baseline, the OA-DG method has higher correlations between the same class, but lower correlations with other classes. 

\section{Conclusion}
In this study, we propose object-aware domain generalization (OA-DG) method for single-domain generalization in object detection. OA-Mix generates multi-domain with mixing strategies and preserved semantics. OA-Loss trains the domain-invariant representations for foreground and background instances from multi-domain. Our experimental results demonstrate that the proposed method can improve the robustness of object detectors to unseen domains.

\section{Acknowledgements}
This work was supported in part by Institute of Information$\And$communications Technology Planning$\And$Evaluation (IITP) grant funded by Korea government (MSIT) (No.2020-0-00440, Development of Artificial Intelligence Technology that Continuously Improves Itself as the Situation Changes in the Real World). This work was supported in part by Korea Evaluation Institute Of Industrial Technology (KEIT) grant funded by the Korea government (MOTIE) (No.20023455, Development of Cooperate Mapping, Environment Recognition and Autonomous Driving Technology for Multi Mobile Robots Operating in Large-scale Indoor Workspace). The students are supported by the BK21 FOUR from the Ministry of Education (Republic of Korea).

\bibliography{aaai24}

\end{document}